%% file: neurips_2024.tex
\documentclass{article}

\usepackage[final]{neurips_2024}

\usepackage{soul}
\usepackage[utf8]{inputenc} % allow utf-8 input
\usepackage[T1]{fontenc}    % use 8-bit T1 fonts
\usepackage{hyperref}       % hyperlinks
\usepackage{url}            % simple URL typesetting
\usepackage{booktabs}       % professional-quality tables
\usepackage{amsfonts}       % blackboard math symbols
\usepackage{nicefrac}       % compact symbols for 1/2, etc.
\usepackage{microtype}      % microtypography
\usepackage{xcolor}         % colors
\usepackage{bm}
\usepackage{subcaption}
\usepackage{svg}
\usepackage{wrapfig}
\usepackage{microtype}
\usepackage{adjustbox}
\usepackage{graphicx}
\usepackage{multirow}
\usepackage{lipsum}
\usepackage{bbm}

\input{math_commands.tex}

\usepackage{amsmath}
\usepackage{amssymb}
\usepackage{mathtools}
\usepackage{amsthm}

\usepackage[capitalize,noabbrev]{cleveref}

\theoremstyle{plain}

\theoremstyle{definition}

\theoremstyle{remark}

\usepackage[textsize=tiny]{todonotes}
\title{Interpretable Mesomorphic Neural Networks \\ For Tabular Data}

\author{%
  Arlind Kadra \\
  Department of Representation Learning\\
  University of Freiburg\\
  \texttt{kadraa@cs.uni-freiburg.de} \\
  \And 
  Sebastian Pineda Arango \\
  Department of Representation Learning\\
  University of Freiburg\\
  \texttt{pineda@cs.uni-freiburg.de} \\
  \And
  Josif Grabocka \\
  Department of Machine Learning\\
  University of Technology Nuremberg\\
  \texttt{josif.grabocka@utn.de} \\
}

\begin{document}

\maketitle

\input{Text/0-Abstract}
\input{Text/1-Introduction}

\input{Text/4-Proposed-Method}
%\input{Text/5 - Motivation}
\input{Text/2-Related-Work}
\input{Text/6-Experimental-Protocol}
\input{Text/7-Experiments}

\input{Text/9-Conclusion}

\input{Text/10-Limitations}
\input{Text/11-Acknowledge}

\bibliography{inn_icml24}
\bibliographystyle{icml2024}

\newpage
\onecolumn
\appendix

\input{Text/appendix-text}
\input{Text/appendix-plots}
\input{Text/appendix-tables}

\clearpage
\newpage

\section*{NeurIPS Paper Checklist}

\begin{enumerate}

\item {\bf Claims}
    \item[] Question: Do the main claims made in the abstract and introduction accurately reflect the paper's contributions and scope?
    \item[] Answer: \answerYes{} % Replace by \answerYes{}, \answerNo{}, or \answerNA{}.
    \item[] Justification: The results in Section~\ref{subsec:motivation} and Section~\ref{sec:experimental_results} validate our claims.

\item {\bf Limitations}
    \item[] Question: Does the paper discuss the limitations of the work performed by the authors?
    \item[] Answer: \answerYes{} % Replace by \answerYes{}, \answerNo{}, or \answerNA{}.
    \item[] Justification: The limitations of the proposed method are mentioned in Section~\ref{sec:limitations}.

\item {\bf Theory Assumptions and Proofs}
    \item[] Question: For each theoretical result, does the paper provide the full set of assumptions and a complete (and correct) proof?
    \item[] Answer: \answerNA{} % Replace by \answerYes{}, \answerNo{}, or \answerNA{}.
    \item[] Justification: There is no theory in this paper.


    \item {\bf Experimental Result Reproducibility}
    \item[] Question: Does the paper fully disclose all the information needed to reproduce the main experimental results of the paper to the extent that it affects the main claims and/or conclusions of the paper (regardless of whether the code and data are provided or not)?
    \item[] Answer: \answerYes{} % Replace by \answerYes{}, \answerNo{}, or \answerNA{}.
    \item[] Justification: In Sections~\ref{sec:proposed_method} and \ref{sec:protocol} we provide all the information regarding our method/baselines and the preprocessing of the data. We additionally open-source the code. Lastly, the results are reproducible as the experiments were seeded.

\item {\bf Open access to data and code}
    \item[] Question: Does the paper provide open access to the data and code, with sufficient instructions to faithfully reproduce the main experimental results, as described in supplemental material?
    \item[] Answer: \answerYes{} % Replace by \answerYes{}, \answerNo{}, or \answerNA{}.
    \item[] Justification: The code is open-sourced (Section~\ref{sec:protocol}) and all of the necessary information regarding the datasets is provided in Table~\ref{app:dataset_info}, combined with their online identifier where they can be easily accessed from OpenML.

\item {\bf Experimental Setting/Details}
    \item[] Question: Does the paper specify all the training and test details (e.g., data splits, hyperparameters, how they were chosen, type of optimizer, etc.) necessary to understand the results?
    \item[] Answer: \answerYes{} % Replace by \answerYes{}, \answerNo{}, or \answerNA{}.
    \item[] Justification: The information is provided in Section~\ref{sec:protocol}. The code is additionally open-sourced.

\item {\bf Experiment Statistical Significance}
    \item[] Question: Does the paper report error bars suitably and correctly defined or other appropriate information about the statistical significance of the experiments?
    \item[] Answer: \answerYes{}{} % Replace by \answerYes{}, \answerNo{}, or \answerNA{}.
    \item[] Justification: Critical difference diagrams that provide statistical significance information are provided in Section~\ref{sec:experimental_results}.
    \item[] Guidelines:
    \begin{itemize}
        \item The answer NA means that the paper does not include experiments.
        \item The authors should answer "Yes" if the results are accompanied by error bars, confidence intervals, or statistical significance tests, at least for the experiments that support the main claims of the paper.
        \item The factors of variability that the error bars are capturing should be clearly stated (for example, train/test split, initialization, random drawing of some parameter, or overall run with given experimental conditions).
        \item The method for calculating the error bars should be explained (closed form formula, call to a library function, bootstrap, etc.)
        \item The assumptions made should be given (e.g., Normally distributed errors).
        \item It should be clear whether the error bar is the standard deviation or the standard error of the mean.
        \item It is OK to report 1-sigma error bars, but one should state it. The authors should preferably report a 2-sigma error bar then state that they have a 96\% CI, if the hypothesis of Normality of errors is not verified.
        \item For asymmetric distributions, the authors should be careful not to show in tables or figures symmetric error bars that would yield results that are out of range (e.g. negative error rates).
        \item If error bars are reported in tables or plots, The authors should explain in the text how they were calculated and reference the corresponding figures or tables in the text.
    \end{itemize}

\item {\bf Experiments Compute Resources}
    \item[] Question: For each experiment, does the paper provide sufficient information on the computer resources (type of compute workers, memory, time of execution) needed to reproduce the experiments?
    \item[] Answer: \answerYes{} % Replace by \answerYes{}, \answerNo{}, or \answerNA{}.
    \item[] Justification: All the necessary information is provided in Section~\ref{sec:protocol} and \ref{sec:experimental_results}.
    
\item {\bf Code Of Ethics}
    \item[] Question: Does the research conducted in the paper conform, in every respect, with the NeurIPS Code of Ethics \url{https://neurips.cc/public/EthicsGuidelines}?
    \item[] Answer: \answerYes{} % Replace by \answerYes{}, \answerNo{}, or \answerNA{}.
    \item[] Justification: The research conducted conforms with the NeurIPS Code of Ethics.

\item {\bf Broader Impacts}
    \item[] Question: Does the paper discuss both potential positive societal impacts and negative societal impacts of the work performed?
    \item[] Answer: \answerYes{} % Replace by \answerYes{}, \answerNo{}, or \answerNA{}.
    \item[] Justification: The impact of our work has been mentioned in the Introduction Section and in Section~\ref{sec:experimental_results}.
    
    
\item {\bf Safeguards}
    \item[] Question: Does the paper describe safeguards that have been put in place for responsible release of data or models that have a high risk for misuse (e.g., pretrained language models, image generators, or scraped datasets)?
    \item[] Answer: \answerNA{} % Replace by \answerYes{}, \answerNo{}, or \answerNA{}.
    \item[] Justification: The paper poses no risks.

\item {\bf Licenses for existing assets}
    \item[] Question: Are the creators or original owners of assets (e.g., code, data, models), used in the paper, properly credited and are the license and terms of use explicitly mentioned and properly respected?
    \item[] Answer: \answerYes{} % Replace by \answerYes{}, \answerNo{}, or \answerNA{}.
    \item[] Justification: Everything that was used from previous work be it a method or dataset has been properly cited in the manuscript.

\item {\bf New Assets}
    \item[] Question: Are new assets introduced in the paper well documented and is the documentation provided alongside the assets?
    \item[] Answer: \answerYes{} % Replace by \answerYes{}, \answerNo{}, or \answerNA{}.
    \item[] Justification: The code is provided and documented.

\item {\bf Crowdsourcing and Research with Human Subjects}
    \item[] Question: For crowdsourcing experiments and research with human subjects, does the paper include the full text of instructions given to participants and screenshots, if applicable, as well as details about compensation (if any)? 
    \item[] Answer: \answerNA{}{} % Replace by \answerYes{}, \answerNo{}, or \answerNA{}.
    \item[] Justification: Not applicable.

\item {\bf Institutional Review Board (IRB) Approvals or Equivalent for Research with Human Subjects}
    \item[] Question: Does the paper describe potential risks incurred by study participants, whether such risks were disclosed to the subjects, and whether Institutional Review Board (IRB) approvals (or an equivalent approval/review based on the requirements of your country or institution) were obtained?
    \item[] Answer: \answerNA{} % Replace by \answerYes{}, \answerNo{}, or \answerNA{}.
    \item[] Justification: Not applicable.

\end{enumerate}

\end{document}

%% file: math_commands.tex
\usepackage{amsmath,amsfonts,bm}

\def\eqref#1{equation~\ref{#1}}
\def\1{\bm{1}}

\DeclareMathAlphabet{\mathsfit}{\encodingdefault}{\sfdefault}{m}{sl}
\SetMathAlphabet{\mathsfit}{bold}{\encodingdefault}{\sfdefault}{bx}{n}

\newcommand{\Ls}{\mathcal{L}}
\newcommand{\R}{\mathbb{R}}

\DeclareMathOperator*{\argmin}{arg\,min}

%% file: Text/0-Abstract.tex
\begin{abstract}

Even though neural networks have been long deployed in applications involving tabular data, still existing neural architectures are not explainable by design. In this work, we propose a new class of interpretable neural networks for tabular data that are both deep and linear at the same time (i.e. mesomorphic). We optimize deep hypernetworks to generate explainable linear models on a per-instance basis. As a result, our models retain the accuracy of black-box deep networks while offering \textit{free lunch} explainability for tabular data by design. Through extensive experiments, we demonstrate that our explainable deep networks %are as accurate as 
have comparable performance to state-of-the-art classifiers on tabular data and outperform current existing methods that are explainable by design. %On the other hand, we showcase the interpretability of our method on a recent benchmark by empirically comparing prediction explainers. The experimental results reveal that our models are comparable in accuracy to their black-box deep-learning counterparts but also as interpretable as state-of-the-art explanation techniques.
\end{abstract}

%% file: Text/1-Introduction.tex
\section{Introduction}

Tabular data are arguably the most widely spread traditional data modality arising in a plethora of real-world application domains~\citep{bischl2021openml,9998482}. There exists a recent trend to deploy neural networks for predictive tasks on tabular data~\citep{kadra2021well, gorishniy2021revisiting,somepalli2022saint, hollmann2023tabpfn}. In a series of such application realms, it is important to be able to explain the predictions of deep learning models to humans~\citep{RasXGD22}, especially when interacting with human decision-makers, such as in healthcare~\citep{gulum2021review,9233366}, or the financial sector~\citep{10.1093/jjfinec/nbaa025}. Heavily parametrized models such as deep neural networks can fit complex interactions in tabular datasets and achieve high predictive accuracy, however, they are not explainable. In that context, achieving both high predictive accuracy and explainability remains an open research question for the Machine Learning community.

%Linear classification models such as logistic or softmax regression are explainable by design~\citep{burkart2021survey}, however, they typically underfit the complex non-linear interaction between the features and the target variable. On the other hand, heavily parametrized models such as deep neural networks can fit complex interactions, however, they are not explainable. 
%How can we explain predictions to a human, if the target variable can only be estimated by very deep and non-linear interactions of the features? Currently, achieving high predictive accuracy and explainability simultaneously presents a contradiction in objectives and remains an open research question.

In this work, we introduce \textit{mesomorphic} neural architectures\footnote{The etymology of the term \textit{mesomorphic} is inspired by Chemistry as "pertaining to an intermediate phase of matter". For instance, a liquid crystal qualifies as both solid and liquid.}, a new class of deep models that are \textbf{both deep and locally linear} at the same time, therefore, offering interpretability by design. In a nutshell, we propose a new architecture that is simultaneously (i) \textit{deep and accurate}, as well as (ii) \textit{linear and explainable} on a per-instance basis. Technically speaking, we learn \textit{deep} hypernetworks that generate \textit{linear} models that are accurate concerning the data point we are interested in explaining. 

%Our method leverages deep neural networks to capture complex feature interactions and achieves better accuracy compared to other explainable methods by design. However, despite being complex and deep, our neural networks offer explainable predictions in the form of simple linear models. What makes this paradigm shift possible is rethinking the role of a neural network in classical supervised learning. Instead of estimating the target variable, we train deep networks that generate the best linear classifier for a data point. 

Our interpretable mesomorphic networks for tabular data (dubbed IMN) are classification or regression models that identify the relevant tabular features by design. It is important to highlight that this work tackles explaining predictions for a single data point~\citep{LundbergL17}, instead of explaining a model globally for the whole dataset~\citep{RasXGD22}. Similarly to existing prior works~\citep{10.5555/3327757.3327875,pmlr-v80-chen18j}, we train deep models that generate explainable local models for a data sample of interest. In contrast, we train hypernetworks that generate linear models in the original feature space through a purely supervised end-to-end optimization. 

We empirically show that the proposed explainable deep models are both as accurate as existing black-box classifiers for tabular datasets and achieve better performance compared to explainable end-to-end prior methods. At the same time, IMN is as interpretable as explainer techniques. Throughout this work, explainers can be categorized into two groups: \textit{i)} interpretable surrogate models that are trained to approximate black-box models~\citep{LundbergL17}, and \textit{ii)} end-to-end explainable methods by design. Concretely, we show that our method achieves comparable accuracy to competitive black-box classifiers and manages to outperform current state-of-the-art end-to-end explainable methods on the tabular datasets of the popular AutoML benchmark~\citep{amlb2019}. In addition, we compare our technique against state-of-the-art predictive explainers on the recent XAI explainability benchmark for tabular data~\citep{xai-bench-2021} and empirically demonstrate that our method offers competitive interpretability. As a result, our method represents a significant step forward in making deep learning explainable by design for tabular datasets. Overall, this work offers the following contributions:

\begin{itemize}
    \item We present a technique that makes deep learning explainable by design via training hypernetworks to generate instance-specific linear models.
    \item We offer ample empirical evidence that our method is as accurate as black-box classifiers, with the benefit of being as interpretable as state-of-the-art prediction explainers.
\end{itemize}

%% file: Text/4-Proposed-Method.tex
\section{Proposed Method}
\label{sec:proposed_method}

\subsection{Shallow Interpretability through Deep Hypernetworks} 
\label{sec:shallowdeep}

Let us denote a tabular dataset consisting of $N$ instances of $M$-dimensional features as $X \in \mathbb{R}^{N \times M}$ and the $C$-dimensional categorical target variable as $Y \in \left\{1,\dots, C\right\}^N$. A model with parameters $w \in \mathcal{W}$ estimates the target variable as $f: \R^M~\times~\mathcal{W}~\rightarrow~\mathbb{R}^C$ and is optimized by minimizing the empirical risk $\argmin_{w \in \mathcal{W}} \sum_{n=1}^N \mathcal{L}\left(y_n, f(x_n;w)\right)$, where $\mathcal{L}: \left\{1,\dots, C\right\} \times \mathbb{R}^C \rightarrow \mathbb{R}_{+}$ is a loss function. An explainable model $f$ is one whose predictions $\hat y_n = f(x_n; w)$ for a data point $x_n$ are interpretable by humans. For instance, linear models and decision trees are commonly accepted to be interpretable by Machine Learning practitioners~\citep{10.1145/2939672.2939778, LundbergL17}.

In this work, we rethink shallow interpretable models $f(x_n; w)$ by defining their parameters $w \in \mathcal{W}$ to be the output of deep non-interpretable hypernetworks 
$w(x_n; \theta): \R^M \times \Theta \rightarrow \mathcal{W}$, 
%$w(x_n; \theta) ; w:  \R^M \times \Theta \rightarrow \mathcal{W}$
where the parameters of the hypernetwork are $\theta \in \Theta$. We remind the reader that a hypernetwork (a.k.a. meta-network, or "network of networks") is a neural network that generates the parameters of another network~\citep{ha2017hypernetworks}. In this mechanism, we train deep non-interpretable hypernetworks to generate interpretable models $f$ in an end-to-end manner as $\argmin_{\theta \in \Theta} \sum_{n=1}^N \mathcal{L}\left(y_n, f(x_n; w(x_n; \theta))\right)$.

%There exists a general consensus that linear classifiers, such as the multinomial logistic regression are explainable by design~\citep{burkart2021survey}. Specifically, an explainable linear classifier with weights $w \in \R^{C\times (M+1)}$ is:

%\vspace{-0.5cm}

%\begin{align}
%\label{eq:softmaxregression}
%    f\left(x_n; w\right)_c &:= \frac{e^{z_c(x_n; w_c)}}{\sum_{k=1}^C e^{z_k(x_n; w_k)}} \\ \nonumber
%    z_c(x_n; w_c) &:= w_c^T x_n + w_{c,0}
%\end{align}

%\begin{align}
%\label{eq:softmaxregression}
%    f\left(x_n; w\right)_c &= 
%    \begin{cases}
%    \frac{e^{z_c}}{\sum_{k=1}^C e^{z_k}}, \;\;\; \text{Softmax Regression} \\ 
%    \frac{1}{1 + e^{{-z}_c}}, \;\;\; \text{Logistic Regression} \\
%    z_c, \;\;\; \text{Linear Regression} \\
%    \end{cases} \nonumber \\
%    \text{ where } \;z_c &:= w_c^T x_n + w_{c,0}\;\;\\ \text{and}\;\; w &\in \R^{C\times (M+1)}, \; \forall c \in \left\{1,\dots, C\right\} \nonumber
%\end{align}

%where $w_{c} \in \R^M$ denotes the weights of the linear model for the $c$-the class, $w_{c,0}$ the corresponding bias term, and $z_c$ the logit predictions. Therefore, it is possible to explain the classification $\hat y_{n, c} = f(x_n; w)_c$ for the $c$-th target class, by analysing the feature attributions $w_{c,m} \, x_{n, m}$.

\subsection{Interpretable Mesomorphic Networks (IMN)}

Our method trains deep Multi-Layer Perceptron (MLP) hypernetworks that generate the parameters of linear models. For the case of multi-class classification, we consider linear models with parameters $w \in \R^{C \times \left(M + 1\right)}$, denoting one set of weights and bias terms per class, as $f\left(x_n; w\right)_c = 
    e^{z\left(x_n; w\right)_c} / \sum_{k=1}^C e^{z\left(x_n; w\right)_k}$, with $z\left(x_n; w\right)_c = \sum_{m=1}^M w_{c,m} x_{n,m} + w_{c,0}$ representing the logit predictions for the $c$-th class. For the case of regression the linear model is simply $f\left(x_n; w\right) = \sum_{m=1}^M w_{m} x_{n,m} + w_{0}$ with $w \in \mathbb{R}^{M+1}$.

Let us present our method IMN by starting with the case of multi-class classification following the hypernetwork mechanism explained in Section~\ref{sec:shallowdeep}. The  hypernetwork $w(x_n; \theta): \R^M \times \Theta \rightarrow \R^{C\times (M+1)}$ with parameters $\theta \in \Theta$ is a function that given a data point $x_n \in \R^M$ generates the predictions as: 

\begin{small}
\begin{align}
    f\left(x_n; w(x_n; \theta)\right)_c = 
    \frac{e^{z\left(x_n; w(x_n; \theta)\right)_c}}{\sum_{k=1}^C e^{z\left(x_n; w(x_n; \theta)\right)_k}}, \;\; z\left(x_n; w(x_n; \theta)\right)_c = \sum_{m=1}^M w\left(x_n; \theta\right)_{c,m} x_{n,m} + w\left(x_n; \theta\right)_{c,0}
\end{align}
\end{small}

Instead of training weights $w$ as in a standard linear classification, we use the output of an MLP network as the linear weights $w(x_n; \theta)$. We illustrate the architecture of our mesomorphic network in Figure~\ref{fig:method_illustration}. In the case of regression, our linear model with hypernetworks is $f(x_n; w(x_n; \theta)) = \sum_{m=1}^M w\left(x_n; \theta\right)_{m} x_{n,m} + w\left(x_n; \theta\right)_{0}$. We highlight that our experimental protocol (Section~\ref{sec:protocol}) includes both classification and regression datasets.

Ultimately, we train the optimal parameters of the hypernetwork to minimize the following loss in an end-to-end manner: $\argmin\limits_{\theta \in \Theta} \sum_{n=1}^N \Ls\left(y_n, f\left(x_n; w(x_n; \theta)\right)\right)  + \lambda ||w\left(x_n; \theta\right)||_1$.

\input{Plots/method_illustration}

Our hypernetworks generate interpretable models that are accurate concerning a data point of interest (e.g. "Explain why patient $x_n$ is estimated to have cancer $f(x_n; w(x_n; \theta)) > 0.5$ by analyzing the impact of features using the generated linear weights."). We stress that our novel method IMN does not simply train one linear model per data point, contrary to prior work~\citep{10.1145/2939672.2939778}. Instead, the hypernetwork learns to generate accurate linear models by a shared network across all data points. As a result, generating the linear weights demands a single forward pass through the hypernetwork, rather than a separate optimization procedure. Furthermore, our method intrinsically learns to generate similar linear hyperplanes for neighboring data instances. The outputted linear models are accurate both in correctly classifying the data point $x_n$, but also for the other majority of training instances in the neighborhood (see proof-of-concept experiment below). The outcome is a linear model with parameters $w(x_n; \theta)$ that both interprets the prediction, but also serves as an accurate local model for the neighborhood of points.

%Therefore, we learn to generate a local linear model $w(x_n; \theta)$ for each instance $x_n$, by minimizing the loss in predicting the target $y_n$. Overall, we define $g$ to be a deep neural network $g: \R^M \rightarrow \R^{C\times(M+1)}$ with $M$ input units and $C\times(M+1)$ output units as illustrated in Figure~\ref{fig:method_illustration}. Furthermore, the hyperparameter $\lambda \in \R_+$ controls the degree of L1 regularization penalty on the weights. The full architecture is trained with standard stochastic gradient descent and backpropagation.

\subsection{Explainability Through Feature Attribution}
\label{subsec:feature_impacts}
The generated linear models $w\left(x_n; \theta\right)$ can be used to explain predictions through feature attribution (i.e. feature importance)~\citep{xai-bench-2021}. It is important to re-emphasize that our method offers interpretable predictions for the estimated target $f(x_n; w\left(x_n; \theta\right))$ of a particular data point $x_n$. Concretely, we can analyse the linear coefficients $\left\{ w(x_n; \theta)_1, \dots,  w(x_n; \theta)_M\right\}$ to distill the importances of $\left\{x_{n, 1},\dots,x_{n, M}\right\}$ by measuring the residual impact on the target. The impact of the $m$-th feature $x_{n, m}$ in estimating the target variable, is proportional to the change in the estimated target if we remove the feature~\citep{10.5555/3454287.3455160}. Considering our linear models, the impact of the $m$-th feature is proportional to the change of the predicted target if we set the $m$-th feature to zero. In terms of notation, we multiply the feature vector 
 element-wise with a Kronecker delta vector $\delta^{m}_i = \mathbbm{1}_{m \ne i}$.

%f(\left\{x_{n, 1},\dots,x_{n, M}\right\}; \theta) - f(\left\{x_{n, 1},\dots, x_{n, m-1}, x_{n, m+1},\dots, x_{n, M}\right\}; \theta ) \propto  \hat w(x_n; \theta)_m \, x_{n, m}$
\vspace{-0.5cm}

\begin{align}
\label{eq:feature_attribution_equation}
f(x_n; w\left(x_n; \theta\right)) -f(x_n \odot \delta^{m}; w\left(x_n; \theta\right) ) \propto  w(x_n; \theta)_m \, x_{n, m}
\end{align}

As a result, our feature attribution strategy is that the $m$-th feature impacts the prediction of the target variable by a signed magnitude of $ w(x_n; \theta)_m \, x_{n, m}$. In our experiments, all the features are normalized to the same mean and variance, therefore, the magnitude $w(x_n; \theta)_m \, x_{n, m}$ can be directly used to explain the impact of the $m$-th feature. In cases where the unsigned importance is required, a practitioner can use the absolute impact $| w(x_n; \theta)_m \, x_{n, m}|$ as the attribution. Furthermore, to measure the global importance of the $m$-th feature for the whole dataset, we can compute $\frac{1}{N}\sum_{n=1}^N | w(x_n; \theta)_m \, x_{n,m}|$.

\subsection{Proof-of-concept: Globally Accurate and Locally Interpretable Classifiers}
\label{subsec:motivation}

\input{Plots/motivation}

As a proof of concept, we run our method on the half-moon toy task that consists of a 2-dimensional tabular dataset in the form of two half-moons that are not linearly separable. 

Initially, we investigate the global accuracy of our method. As shown in Figure~\ref{fig:motivation} (left), our method correctly classifies all the examples. Furthermore, our method learns an optimal non-linear decision boundary that separates the classes (plotted in green). To determine the decision boundary, we perform a fine-grid prediction on all possible combinations of $x_1$ and $x_2$. Subsequently, we identify the points that exhibit the minimal prediction distance to a probability prediction of 0.5. Lastly, in Figure~\ref{fig:motivation} (right) we investigate the local interpretability of our method, by taking a point $x'$ and calculating the corresponding weights $\left({w}\left(x'\right), {w}\left(x'\right)_0\right)$ generated by our hypernetwork, where we omited the dependence on $\theta$ for simplicity. The black line shows all the points that reside on the hyperplane $ w(x')$ as $\{x\,|\,{w}\left(x'\right)^T\,x + {w}_0\left(x'\right) = 0\}$. It is important to highlight that the local hyperplane does not only correctly classify the point $x'$, but also the neighboring points, retaining an accurate linear classifier for the neighborhood of points. 

\input{Tables/local_hyperplane_accuracy}

To validate our claim that the per-example (local) hyperplane correctly classifies neighboring points, we conduct the following analysis: For every datapoint $x_n$ we take a specific number of nearest neighbor examples from every class, and we evaluate the classification accuracy of the hyperplane generated for the datapoint $x_n$ on the set of all neighbors. We repeat the above procedure with varying neighborhood sizes and we present the results in Table~\ref{table:local_hyperplane_accuracy}. The results indicate that the mesomorphic neural network generates hyperplanes that are accurate in the neighborhood of the point whose prediction we are interested in explaining.

%Lastly, we investigate the resilience of our method in finding the important features in the presence of noisy features. We add from 2 up to 8 additional random features to the dataset and investigate the weights $\left(|\hat{w}(x)_1| / \sum_k |\hat{w}(x)_k|, |\hat{w}(x)_2| / \sum_k |\hat{w}(x)_k|\right)$. The results in Figure~\ref{fig:motivation} (\textbf{Right}) demonstrate that the original features $\left(x_1, x_2\right)$ retain their high linear weight values despite the presence of noisy features. %In more detail, $|\hat{w}_2|$ is not changed, while, $|\hat{w}_1|$ slightly drops by 0.03 while, still retaining the majority of the weight importance after $|\hat{w}_2|$.

%% file: Plots/method_illustration.tex
\begin{wrapfigure}{!hr}{0.5\textwidth}
    \centering
    \includegraphics[width=0.5\textwidth]{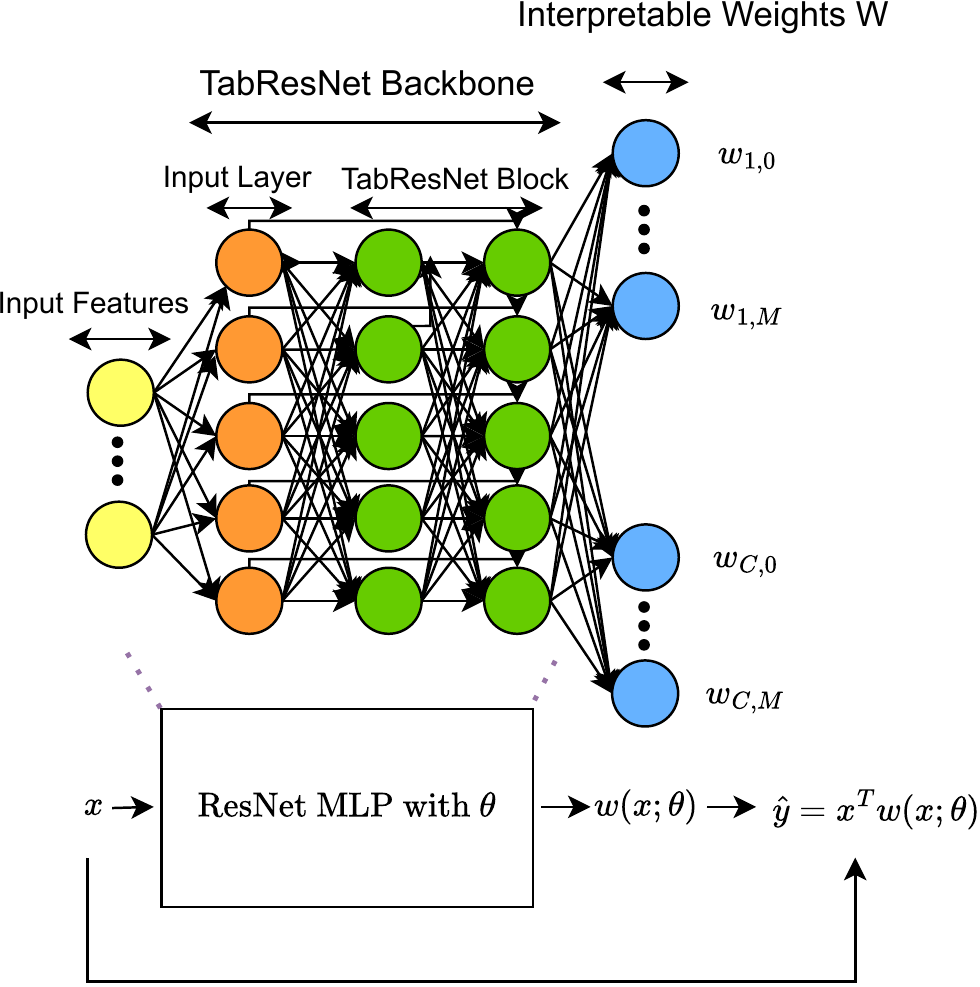}
    \caption[Method Illustration]{The IMN architecture.} 
    \label{fig:method_illustration}
    \vspace{-0.4cm}
\end{wrapfigure}

%% file: Plots/motivation.tex
\begin{figure*}
\centering
\includegraphics[width=0.9\textwidth]{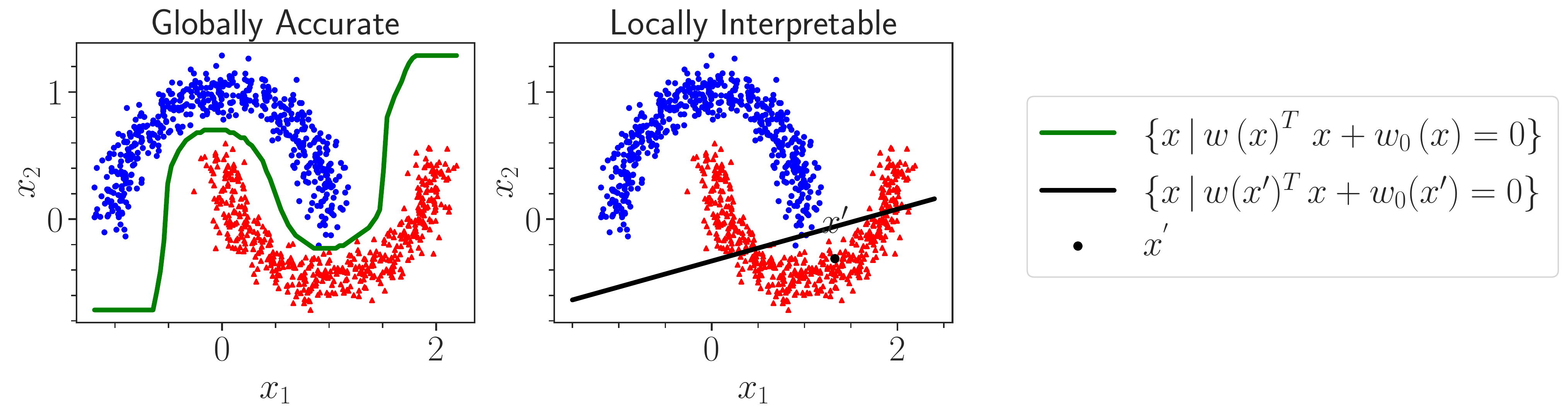}
    \caption[Interpretability of our Neural Network.]{Investigating the accuracy and interpretability of IMN. \textbf{Left:} The global decision boundary of our method that separates the classes correctly. \textbf{Right:} The local hyperplane pertaining to an example $x'$ which correctly classifies the local example and retains a good global classification for the neighboring points.}% \textbf{Right:} The resilience to noise by retaining the weight importance of the original features in the presence of an incremental number of random features.}
    \label{fig:motivation}
    %\vspace{-0.5cm}
\end{figure*}

%% file: Tables/local_hyperplane_accuracy.tex
\begin{wraptable}{!tr}{0.4\textwidth}
%\begin{table}
\caption{Accuracy of local hyperplanes for neighboring points.}
\label{table:local_hyperplane_accuracy}
\centering
\footnotesize
\begin{adjustbox}{width=0.3\textwidth}{
\begin{tabular}{@{}cc@{}}
 \toprule
 \textbf{Number of Neighbors} & \textbf{Accuracy} \\
 \midrule
  10 & 0.84 \\
  25 & 0.82 \\
  50 & 0.78 \\
  100 & 0.77 \\
  200 & 0.77 \\
  
 \bottomrule 
\end{tabular}
}\end{adjustbox}
%\vspace{-0.4cm}
%\end{table}
\end{wraptable}

%% file: Text/2-Related-Work.tex
\section{Related Work}
\label{sec:relatedwork}

%A similar work~\citep{agarwal2021neural} that is trained in an end-to-end manner, proposes to learn an additive function by using a neural network in isolation for every feature to model its contribution to the output. However, it does not consider interactions between features like INN. Additionally, it has a higher complexity in the number of parameters since every feature is associated with its dedicated neural network. A complexity that is further increased in multiclass problems.

\paragraph{Interpretable Models by Design:} There exist Machine Learning models that offer interpretability by default. A standard approach is to use linear models \citep{tibshirani1996regression, efron2004,berkson1953statistically} that assign interpretable weights to each of the input features. On the other hand, decision trees~\citep{loh2011classification,craven1995extracting} use splitting rules that build up leaves and intermediate nodes. Every leaf node is associated with a predicted label, making it possible to follow the rules that led to a specific prediction. Bayesian methods such as Naive Bayes \citep{murphy2006naive} or Bayesian Neural Networks \citep{friedman1997bayesian} provide a framework for reasoning on the interactions of prior beliefs with evidence, thus simplifying the interpretation of probabilistic outputs. Instance-based models allow experts to reason about predictions based on the similarity to the train samples. The prediction model aggregates the labels of the neighbors in the training set, using the average of the top-k most similar samples~\citep{freitas2014comprehensible,kim2015mind}, or decision functions extracted from prototypes~\cite{MartensBGV07}. Attention-based models like TabNet~\citep{arik2021tabnet} make use of sequential attention to generate feature weights on a per-instance basis, while, DANet~\citep{chen2022danets} generates global importance weights for both the raw input features and higher order concepts. Neural additive models (NAMs)~\citep{agarwal2021neural} use a neural network per feature to model the additive function of individual features to the output. However, these models trade-off the performance for the sake of interpretability, therefore challenging their usage on applications that need high performance. A prior similar work also trains hyper-networks to generate local models by learning prototype instances through an encoder model~\cite{10.5555/3327757.3327875}. In contrast, we directly generate interpretable linear models in the original feature space. %A drawback of NAMs compared to our method is that it considers features in isolation. Additionally, NAMs have a higher complexity in the number of parameters since every feature is associated with its dedicated neural network. A complexity that is further increased in multiclass problems. 

\paragraph{Interpretable Model Distillation:} Given the common understanding that complex models are not interpretable, prior works propose to learn simple surrogates for mimicking the input-output behavior of the complex models~\citep{burkart2021survey}.  Such surrogate models are interpretable, such as linear regression or decision trees~\citep{10.1145/2939672.2939778}. The local surrogates generate interpretations only valid in the neighborhood of the selected samples. Some approaches explain the output by computing the contribution of each attribute~\citep{LundbergL17} to the prediction of the particular sample. An alternative strategy is to fit globally interpretable models, by relying on decision trees~\citep{FrosstH17, Yang0R18}, or linear models~\citep{10.1145/2939672.2939778}. Moreover, global explainers sometimes provide feature importances~\citep{goldstein2015peeking, CortezE11}, which can be used for auxiliary purposes such as feature engineering. Most of the surrogate models tackle the explainability task disjointly, by first training a black box model, then learning a surrogate in a second step. 

\paragraph{Interpretable Deep Learning via Visualization:} 
Given the success of neural networks in real-world applications in computer vision, a series of prior works~\citep{RasXGD22} introduce techniques aiming at explaining their predictions. A direct way to measure the feature importance is by evaluating the partial derivative of the network given the input~\citep{simonyan2013deep}. CAM upscales the output of the last convolutional layers after applying Global Average Pooling (GAP), obtaining a map of the class activations used for interpretability~\citep{ZhouKLOT16}. DeepLift calculates pixel-wise relevance scores by computing differences with respect to a reference image~\citep{ShrikumarGK17}. Integrated Gradients use a baseline image to compute the cumulative sensibility of a black-box model $f$
to pixel-wise changes~\citep{SundararajanTY17}. 
Other methods directly compute the pixel-wise relevance scores such that the network's output equals the sum of scores computed via Taylor Approximations~\citep{MontavonLBSM17}.
%In another stream of works, perturbation-based methods offer interpretability by occluding a group of pixels on the black-box model~\citep{ZeilerF14}, or by adding a small amount of noise to the input of a surrogate model \citep{FongV17}. Alternative approaches propose to compute the relevance of a specific input feature by marginalizing out the prediction of a probabilistic model \citep{ZintgrafCAW17}. %Similar to surrogate methods, the approaches for visually explaining DNN outputs re-use the black-box model as part of an external mechanism to add interpretability, incurring time overheads due to the disjoint learning of the black-box and the explaining agent. 

%% file: Text/6-Experimental-Protocol.tex
\section{Experimental Protocol}
\label{sec:protocol}

\subsection{Predictive Accuracy Experiments}

\paragraph{Baselines:} In terms of interpretable white-box classifiers, we compare against \textbf{Logistic Regression} and \textbf{Decision Trees}, based on their scikit-learn library implementations~\citep{pedregosa2011scikit}. On the other hand, we compare against two strong classifiers on tabular datasets, \textbf{Random Forest} and \textbf{CatBoost}. We use the scikit-learn interface for Random Forest, while for CatBoost we use the official implementation provided by the authors~\citep{prokhorenkova2018catboost}. Lastly, in terms of interpretable deep learning architectures, we compare against \textbf{TabNet}~\citep{arik2021tabnet}, a transformer architecture that makes use of attention for instance-wise feature-selection and \textbf{NAM}~\citep{agarwal2021neural}, a neural additive model which learns an additive function for every feature. For TabNet we use a well-maintained public implementation~\footnote{\url{https://github.com/dreamquark-ai/tabnet}}, while, for NAM we use the official public implementation from the authors~\footnote{\url{https://github.com/AmrMKayid/nam}}.

\paragraph{Protocol:} We run our predictive accuracy experiments on the AutoML benchmark that includes 35 diverse classification problems, containing between 690 and 539\,383 data points, and between 5 and 7\,201 features. For more details about the datasets included in our experiments, we point the reader to Appendix~\ref{appendix:tables}. In our experiments, numerical features are standardized, while we transform categorical features through one-hot encoding. For binary classification datasets we use target encoding, where a category is encoded based on a shrunk estimate of the average target values for the data instances belonging to that category. In the case of missing values, we impute numerical features with zero and categorical features with a new category representing the missing value. For CatBoost and TabNet we do not encode categorical features since the algorithms natively handle them. For all the methods considered we tune the hyperparameters with Optuna~\citep{akiba2019optuna}, a well-known hyperparameter optimization (HPO) library. We use the default HPO algorithm (TPE) from the library and we tune every method for 100 HPO trials or a wall-time limit of 1 day, whichever condition gets fulfilled first. The HPO search spaces of the different baselines were taken from prior work~\citep{gorishniy2021revisiting, hollmann2023tabpfn}. For a more detailed description, we kindly refer the reader to Appendix~\ref{appendix:tables}. Additionally, we use the area under the ROC curve (AUROC) as the evaluation metric. Lastly, the methods that offer GPU support are run on a single NVIDIA RTX2080Ti, while, the rest of the methods are run on an AMD EPYC 7502 32-core processor.

%When reporting results, we average across 10 repetitions for every method considered in the experiments. 

\subsection{Explainability Experiments}
\label{sec:explainabilityexperiments}

\paragraph{Baselines:} First, we compare against \textbf{Random}, a baseline that generates random importance weights. Furthermore, \textbf{BreakDown} decomposes predictions into parts that can be attributed to certain features~\citep{staniak2018explanations}. \textbf{TabNet} offers instance-wise feature importances by making use of attention. \textbf{LIME} is a local interpretability method~\citep{10.1145/2939672.2939778} that fits an explainable surrogate (local model) to single instance predictions of black-box models. On the other hand, \textbf{L2X} is a method that applies instance-wise feature selection via variational approximations of mutual information~\citep{pmlr-v80-chen18j} by making use of a neural network to generate the weights of the explainer. \textbf{MAPLE} is a method that uses local linear modeling by exploring random forests as a feature selection method~\citep{plumb2018model}. \textbf{SHAP} is an additive feature attribution method~\citep{LundbergL17} that allows local interpretation of the data instances. Last but not least, \textbf{Kernel SHAP} offers a reformulation of the LIME constrains~\citep{LundbergL17}.

\paragraph{Metrics and Benchmark:} As explainability evaluation metrics we use faithfulness~\citep{LundbergL17}, monotonicity~\citep{10.1145/3447548.3467265} (including the ROAR variants~\citep{10.5555/3454287.3455160}), infidelity~\citep{10.5555/3454287.3455271} and Shapley correlation~\citep{LundbergL17}. For a detailed description of the metrics, we refer the reader to XAI-Bench, a recent explainability benchmark~\citep{xai-bench-2021}.

For our explainability-related experiments, we use all three datasets (Gaussian Linear, Gaussian Non-Linear, and Gaussian Piecewise) available in the XAI-Bench~\citep{xai-bench-2021}. For the state-of-the-art explainability baselines, we use the Tabular ResNet (TabResNet) backbone as the model for which the predictions are to be interpreted (same as for IMN).
We experiment with different versions of the datasets that feature diverse $\rho$ values, where $\rho$ corresponds to the amount of correlation among features. All datasets have a train/validation set ratio of 10 to 1. 

\paragraph{Implementation Details:} We use PyTorch as the main library for our implementation. As a backbone, we use a TabResNet where the convolutional layers are replaced with fully-connected layers as suggested by recent work~\citep{kadra2021well}. For the default hyperparameters of our method, we use 2 residual blocks and 128 units per layer combined with the GELU activation~\citep{hendrycks2016gaussian}. When training our network, we use snapshot ensembling~\citep{huang2017snapshot} combined with cosine annealing with restarts~\citep{loshchilov2018decoupled}. We use a learning rate and weight decay value of 0.01, where, the learning rate is warmed up to 0.01 for the first 5 epochs, a dropout value of 0.25, and an L1 penalty of 0.1 on the weights. Our network is trained for 500 epochs with a batch size of 64. We make our implementation publicly available\footnote{Source code at \url{https://github.com/ArlindKadra/IMN}}.
%\footnote{Source code at \url{https://anonymous.4open.science/r/IMN/}}.

%% file: Text/7-Experiments.tex
\section{Experiments and Results}
\label{sec:experimental_results}

\paragraph{Hypothesis 1:} IMN outperforms interpretable white-box models in terms of predictive accuracy.

We compare our method against decision trees and logistic regression, two white-box interpretable models. We run all aforementioned methods on the AutoML benchmark and we measure the predictive performance in terms of AUROC. Lastly, we measure the statistical significance of the results using the autorank package~\cite{Herbold2020} that runs a Friedman test with a Nemenyi post-hoc test, and a $0.05$ significance level. Figure~\ref{fig:wb_methods_diagram} presents the average rank across datasets based on the AUROC performance. As observed IMN achieves the best rank across the AutoML benchmark datasets. Furthermore, the difference is statistically significant against both decision trees and logistic regression. The detailed per-dataset results are presented in Appendix~\ref{appendix:tables}.

%Decision trees have the worst rank compared to all white-box models, the difference in performance to Logistic/Softmax regression is statistically significant with a $p$-value of $5e$-$07$. 

%The difference IMN where, the difference is statistically significant with a $p$-value of $2.95e$-$07$. Lastly, IMN outperforms Logistic/Softmax regression with statistically significant different with a $p$-value of $1.8e$-$05$. Based on the results, we conclude that \textbf{IMNs outperform white-box interpretable methods.}

\paragraph{Hypothesis 2:} The explainability of IMN does not have a statistically significant negative impact on predictive accuracy. Additionally, it achieves a comparable performance against state-of-the-art methods.

\input{Plots/white_box_models_cd}

This experiment addresses a simple question: \textit{Is our explainable neural network as accurate as a black-box neural network counterpart, that has the same architecture and same capacity?}. Since our hypernetwork is a slight modification of the TabResNet~\cite{kadra2021well}, we compare it against TabResNet as a classifier. For completeness, we also compare against four other strong baselines, Gradient-Boosted Decision Trees (CatBoost), Random Forest, TabNet, and NAMs. Since the official implementation of NAMs only supports binary classification and regression, we separate the results into: \textit{i}) results over 18 binary classification datasets (Figure~\ref{fig:black_box_cd} \textbf{Top}), and \textit{ii}) results over all datasets (Figure~\ref{fig:black_box_cd} \textbf{Bottom}).

\input{Plots/black_box_models_cd}

%We additionally compare against TabNet, a deep-learning architecture that offers interpretability.

%As in Experiment 1, we compare the methods test performance and we analyze if the differences in the method performances are statistically significant. In the experiment, we do not perform hyperparameter tuning and we use the default configurations for all methods, despite recent work~\citep{kadra2021well} that indicates that well-tuned neural networks can achieve state-of-the-art results against traditional state-of-the-art models in tabular data.

The results of Figure~\ref{fig:black_box_cd} demonstrate that IMN achieves a comparable performance to state-of-the-art tabular classification models, while significantly outperforming explainable methods by design. IMN achieves a comparable performance to TabResNet, while outperforming TabNet and NAMs, indicating that its explainability does not harm accuracy in a significant way. There is no statistical significance of the differences between IMN, TabResNet and CatBoost. However, the difference in performance between IMNs, Random Forest, TabNet and NAMs is statistically significant. 

\input{Tables/method_runtimes}
Additionally, we investigate the runtime performance of the different baselines (NAM is excluded since it cannot be run on the full benchmark). We present the results in Table~\ref{table:method_runtimes}. As expected, deep learning methods take a longer time to train, however, both IMN and TabResNet are the most efficient during inference. We observe that TabResNet takes longer to converge compared to IMN\footnote{The number of training epochs is a hyperparameter.}, however, both methods demand approximately the same inference time. As a result, the explainability of our method comes as a \textit{free-lunch} benefit. Lastly, IMN is \textbf{64x faster in inference} compared to TabNet, an end-to-end deep-learning interpretable method. \textbf{Hypothesis 1 and 2 are valid even when default hyperparameters are used}, for more details we kindly refer the reader to Appendix~\ref{appendix:plots}.

%Furthermore, our method is competitive against state-of-the-art black-box models, such as CatBoost and Random Forest. Although IMN has a worse rank compared to CatBoost, the difference in results is not statistically significant. 

%
%\input{Plots/black_box_models_gain}

%To further investigate the results on individual datasets, in Figure~\ref{fig:black_box_gain} we plot the distribution of the gains in performance of all methods over a single decision tree model. The gain $G$ of a method $m$ run on a dataset $D$ for a single run is calculated as shown in Equation~\ref{eq:gain_calc}.

%\begin{align}
%    G\left(m, D\right) = &\text{AUROC}(m, D) \;/\; \label{eq:gain_calc} \\ &\text{AUROC}(\text{DTree}, D) \nonumber
%\end{align}

%The results indicate that all methods except NAM achieve a comparable gain in performance across the AutoML benchmark datasets, while, the latter achieves a worse performance overall. We present detailed results in Appendix~\ref{appendix:tables}.  

%Another important takeaway is that our reformulation does not degrade the performance of the underlying TabResNet backbone and it has a negligible impact on the runtime, where, the IMN runtime is slower compared to the TabResnet architecture by a factor of only 0.04 $\pm$ 0.03\% across datasets.

%Given the results, we validate that \textbf{IMNs do not degrade the performance of the underlying architecture and have a negligible impact on time, achieving comparable results to state-of-the-art architectures.}

\paragraph{Hypothesis 3:} IMNs offer competitive levels of interpretability compared to state-of-the-art explainer techniques. 

\input{Tables/xaibench}
\input{Plots/xaibench_correlation}

We compare against 8 explainer baselines in terms of 5 explainability metrics in the 3 datasets of the XAI benchmark~\citep{xai-bench-2021}, following the protocol we detailed in Section~\ref{sec:explainabilityexperiments}. The results of Table~\ref{table:xai_interpretability} demonstrate that IMN is competitive against all explainers across the indicated interpretability metrics. We tie in performance with the second-best method Kernel-SHAP~\citep{LundbergL17} and perform strongly against the other explainers. It is worth highlighting that in comparison to all the explainer techniques, the interpretability of our method comes as \textit{a free-lunch}. In contrast, all the rival methods except TabNet are surrogate interpretable models to black-box models. Moreover, IMN strongly outperforms TabNet, the other baseline that offers explainability by design, achieving both better interpretability (Table~\ref{table:xai_interpretability}) and better accuracy (Figure~\ref{fig:black_box_cd}).

\input{Tables/interpretable_method_runtimes}

As a result, for all surrogate interpretable baselines we first need to train a black-box model. Then, for the prediction of \textbf{every} data point, we additionally train a local explainer around that point by predicting with the black-box model multiple times. In stark contrast, our method combines prediction models and explainers as an all-in-one neural network. To generate an explainable model for a data point $x_n$, IMN does not need to train a per-point explainer. Instead, IMN requires only a forward pass through the trained hypernetwork to generate a linear explainer. To quantify the difference in runtime between our method and other interpretable methods we compare the runtimes on a few datasets from the benchmark with a varying number of instances/features such as Credit-g (1000/21), Adult (48842/15), and Christine (5418/1637). Table~\ref{table:interpretable_method_runtimes} presents the results, where, as observed IMN has the fastest inference times, being \textbf{11-65x faster} compared to TabNet which employs attention, \textbf{1710-11400x faster} compared to SHAP that uses the same (TabResNet) backbone, and \textbf{455-215850x faster} compared to SHAP that uses CatBoost as a backbone. 

\input{Tables/census_importances}

Lastly, we compare all interpretability methods on 4 out of 5 metrics in the presence of a varying $\rho$ factor, which controls the correlation of features on the Gaussian Linear dataset. Figure~\ref{fig:xaibench_correlation} presents the comparison, where IMN behaves similarly to other interpretable methods and has a comparable performance with the top methods in the majority of metrics. The results agree with the findings of prior work~\citep{xai-bench-2021}, where the performance in the interpretability metrics drops in the presence of feature correlations. Although our work focuses on tabular data, in Appendix~\ref{appendix:image_application} we present an application of IMN in the vision domain.

\vspace{0.4cm}

\paragraph{Hypothesis 4:} IMN offers a global (dataset-wide) interpretability of feature importance.

The purpose of this experiment is to showcase that IMN can be used to analyze the global interpretability of feature attributions, where the dataset-wide importance of the $m$-th feature is aggregated as $\frac{1}{N}\sum_{n=1}^N |w(x_n; \theta)_m \, x_{n,m}|$. Since we are not aware of a public benchmark offering ground-truth global interpretability of features, we experiment with the Adult Census Income~\citep{kohavi1996scaling}, a very popular dataset, where the goal is to predict whether income exceeds \$50K/yr based on census data. We consider Decision Trees, CatBoost, TabNet, and IMN as explainable methods. Additionally, we use SHAP to explain the predictions of the TabResNet backbone.

\input{Plots/importance_topk}
\vspace{0.4cm}

We present the importance that the different methods assign to features in Table~\ref{table:census_importances}. To verify the feature rankings generated by the models, we analyze the top 5 features of every individual method by investigating the drop in model performance if we remove the feature. The more important a feature is, the more accuracy should drop when removing that feature. The results of Figure~\ref{fig:importance_topk} show that IMNs have a higher relative drop in the model's accuracy when the most important predicted feature is removed. This shows that the feature ranking generated by IMN is proportional to the predictive importance of the feature and monotonously decreasing. In contrast, in the case of CatBoost, TabNet, SHAP, and Decision Trees, the decrease in accuracy is not proportional to the order of the feature importance (e.g. the case of Top-1 for Decision Trees, TabNet, SHAP or Top-2 for CatBoost).

\vspace{0.4cm}

\input{Plots/odor_feature_impacts}

We additionally consider the task of predicting mushroom edibility~\citep{lincoff1997field}. The \textit{odor} feature allows one to predict whether a mushroom is edible or not and basing the predictions only on odor would allow a model to achieve more than 98.5\% accuracy~\citep{arik2021tabnet}. We run IMNs on the mushroom edibility task and we achieve a perfect test AUROC of 1. Furthermore, in Figure~\ref{fig:odor_feature_impacts} we investigate the impact of every feature as described in Section~\ref{subsec:feature_impacts}, where, as observed, our method correctly identifies \textit{odor} as the feature with the highest impact in the output.
Based on the results, we conclude that \textbf{IMNs offer global interpretability}.

%Effectively, we observe that IMNs achieve great explainability, by being able to capture relevant image pixels and keeping up a high prediction performance.} 

%% file: Plots/white_box_models_cd.tex
\begin{wrapfigure}{!hr}{0.5\textwidth}
    \centering
    \includegraphics[width=0.5\textwidth]
    {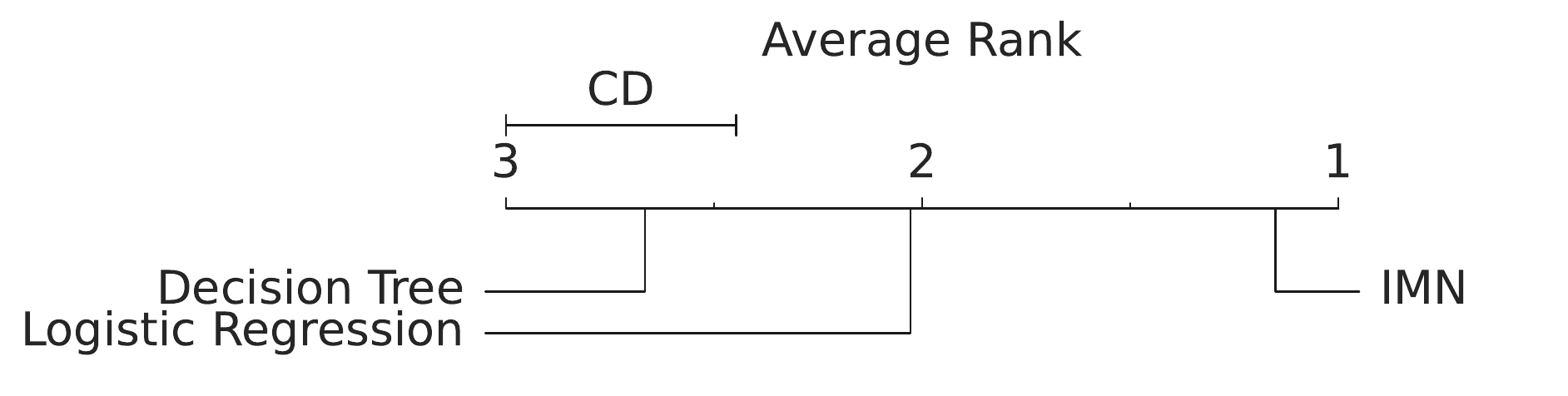}
    \caption[White Box methods diagram.]{The critical difference diagram for the white-box interpretable methods. A lower rank indicates a better performance over datasets.}
    \label{fig:wb_methods_diagram}
\end{wrapfigure}

%% file: Plots/black_box_models_cd.tex
\begin{wrapfigure}{!hr}{0.5\textwidth}
    \centering
    \begin{subfigure}{\textwidth}
    \includegraphics[width=0.45\textwidth]{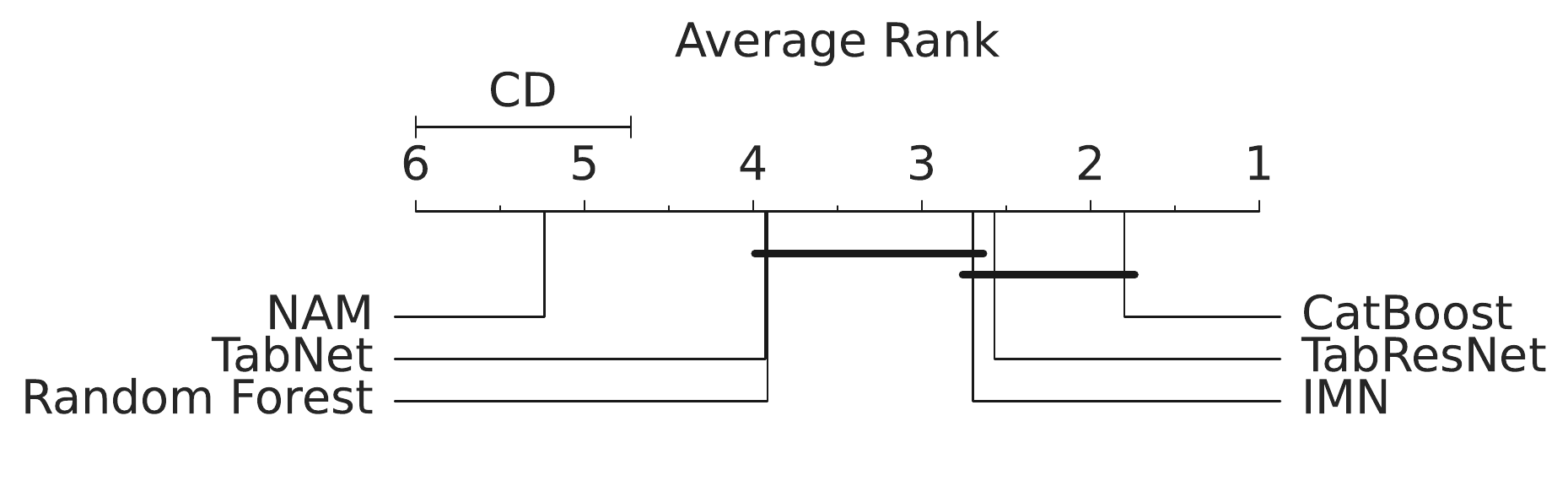}
    \end{subfigure} \hfil % <-- added
    \begin{subfigure}{\textwidth}
    \includegraphics[width=0.45\textwidth]{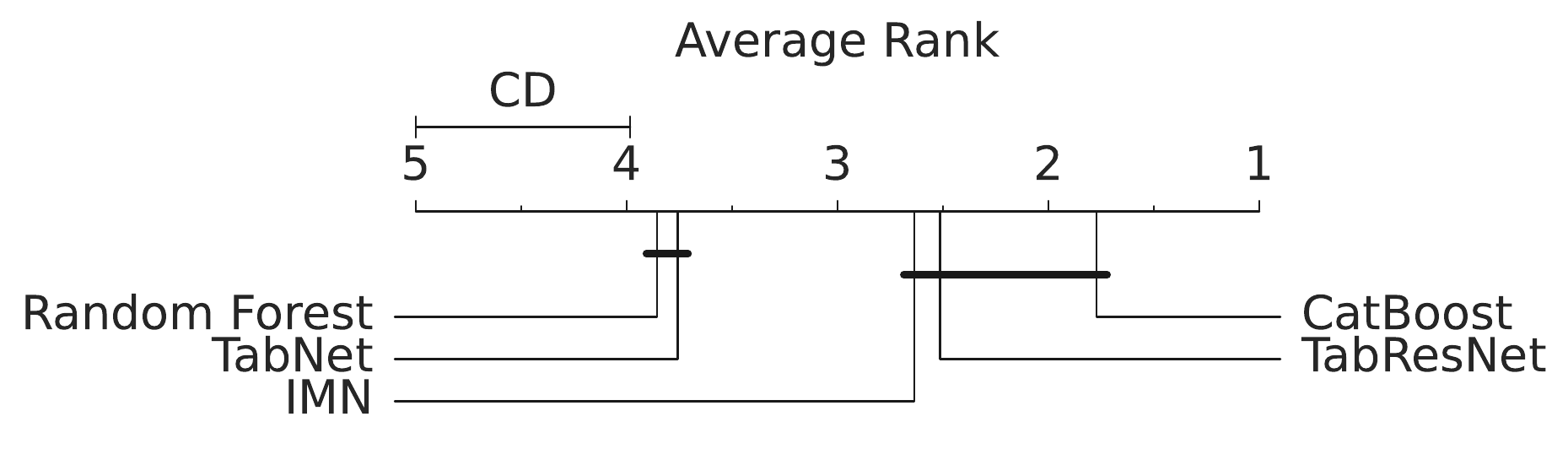}
%\caption{Gain over the number of examples}
%\label{fig:1}
\end{subfigure}%\hfil % <-- added
\caption[Black-box methods comparison]{Black-box methods comparison with critical difference diagrams. \textbf{Top}: The average rank for the binary datasets present in the benchmark. \textbf{Bottom:} The average rank for all datasets present in the benchmark. A lower rank indicates a better performance. Connected ranks via a bold bar indicate that performances are not significantly different ($p > 0.05$).}
\label{fig:black_box_cd}
\vspace{-0.5cm}
\end{wrapfigure}

%% file: Tables/method_runtimes.tex
\begin{wraptable}{!tr}{0.5\textwidth}
%\begin{table}
\caption{Aggregated training and inference times for all methods.}
\label{table:method_runtimes}
\centering
\footnotesize
\begin{adjustbox}{width=0.5\textwidth}{
\begin{tabular}{@{}lcc@{}}
 \toprule
 \textbf{Method Name} & \textbf{Median Training Time (sec)} & \textbf{Median Inference Time (sec)} \\
 \midrule
  IMN (GPU) & 192 & 0.025 \\
  TabResNet (GPU) & 252 & 0.020 \\
  TabNet (GPU) & 237 & 1.60 \\
  CatBoost (GPU) & 63.2 & 0.20 \\
  Random Forest & 42.55 & 2.20 \\
  Logistic Regression & 0.23 & 0.07 \\
  Decision Tree & 0.4 & 0.06 \\
  
 \bottomrule 
\end{tabular}
}\end{adjustbox}
\vspace{-0.3cm}
%\end{table}
\end{wraptable}

%% file: Tables/xaibench.tex
\begin{table*}[!ht]
\caption{Investigating the interpretability of IMNs against state-of-the-art interpretability methods. The results are generated from the XAI Benchmark~\citep{xai-bench-2021} datasets (with $\rho=0$).}
\label{table:xai_interpretability}
\centering
\footnotesize
\begin{adjustbox}{width=1\textwidth}{
\begin{tabular}{@{}llccccccccc@{}}
 \toprule
 \textbf{Metric} & \textbf{Dataset} & \textbf{Random} & \textbf{Breakd.} & \textbf{Maple} & \textbf{LIME} & \textbf{L2X} & \textbf{SHAP} & \textbf{K. SHAP} & \textbf{TabNet} & \textbf{IMN} \\
 \midrule
 \multirow{3}{*}{Faithfulness $\left(\uparrow\right)$} & Gaussian Linear & 0.004 & 0.645 & 0.980 & 0.882 & 0.010 & 0.974 & 0.981 & 0.138 &  \textbf{0.987} \\ 
& Gaussian Non-Linear & -0.079 & -0.001 & 0.487 & 0.796 & 0.155 & 0.926 & \textbf{0.970} & 0.161 & 0.621 \\
& Gaussian Piecewise  & 0.091 & 0.634 & 0.967 & 0.929 & 0.016 & 0.981 & \textbf{0.990} & 0.058 &  0.841 \\ \midrule
\multirow{3}{*}{Faithfulness (ROAR) $\left(\uparrow\right)$} 
& Gaussian Linear & -0.039 & 0.494 & 0.548 & 0.544 & 0.049 & 0.549 & 0.550 & 0.041 & \textbf{0.639}  \\ 
& Gaussian Non-Linear   & 0.050 & 0.006 & 0.040 & -0.040 & -0.060 & -0.010 & -0.036 & -0.001 & 0.027  \\ 
& Gaussian Piecewise & -0.055 & 0.372 & 0.347 & 0.450 & 0.015 & 0.409 & 0.426 & 0.072 & 0.404  \\ \midrule 
 \multirow{3}{*}{Infidelity $\left(\downarrow\right)$} 
   & Gaussian Linear & 0.219 & 0.041 & \textbf{0.007}  & \textbf{0.007} & 0.034 & \textbf{0.007} & \textbf{0.007} & 0.049 &  \textbf{0.007}  \\
  & Gaussian Non-Linear & 0.075 & 0.086 & 0.021 & 0.071 & 0.089 & 0.030  & 0.022 & 0.047 & \textbf{0.018} \\
& Gaussian Piecewise & 0.132 & 0.047 & 0.014  & 0.019 & 0.070 & 0.016  & 0.019 & 0.046 & \textbf{0.008}  \\ \midrule
 \multirow{3}{*}{Monotonicity (ROAR) $\left(\uparrow\right)$} & Gaussian Linear & 0.487 & 0.605 & 0.700 & 0.652 & 0.437 & 0.680 & 0.667 & 0.585 & \textbf{0.785}  \\ 
& Gaussian Non-Linear  & 0.497 & 0.542 & 0.645 & 0.587 & 0.457 & \textbf{0.670} & 0.632 & 0.493 & 0.637  \\ 
& Gaussian Piecewise & 0.485 & 0.665 & 0.787 & 0.427 & 0.442 & 0.717 & \textbf{0.797} & 0.542 & 0.682   \\ \midrule 
\multirow{3}{*}{Shapley Correlation $\left(\uparrow\right)$} 
& Gaussian Linear & -0.016 & 0.246 & \textbf{0.999} & 0.942 & -0.214 & 0.993 & \textbf{0.999} &  0.095 & \textbf{0.999}   \\ 
& Gaussian Non-Linear & -0.069 & -0.179 & 0.686 & 0.872 & -0.095 & 0.974 & \textbf{0.999} & 0.125 & 0.741  \\ 
& Gaussian Piecewise & -0.078 & 0.099 & 0.983 & 0.959 & 0.157 & 0.991 & \textbf{0.999} & 0.070 & 0.875  \\ \midrule
\textbf{Total Wins} &  & 1 & 0 & 2 & 2 & 0 & 2 & \textbf{7} & 0 & \textbf{7} \\
\bottomrule 
\end{tabular}
}\end{adjustbox} 
\end{table*}

%% file: Plots/xaibench_correlation.tex
\begin{figure*}[!ht] 
\includegraphics[width=\textwidth]{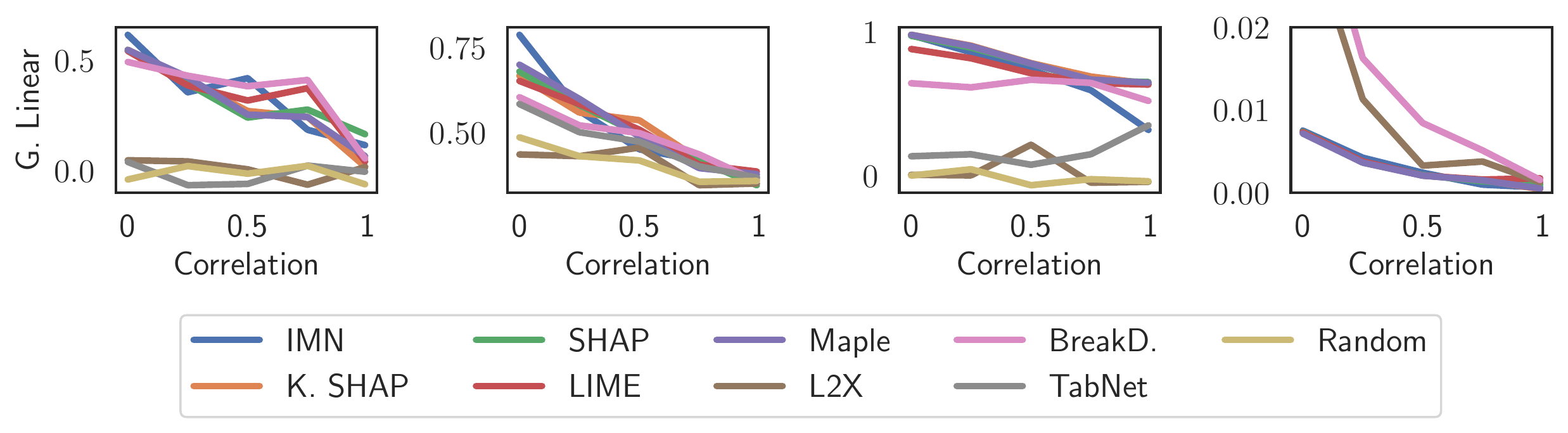}
    \caption[Interpretability analysis over correlation.]{Performance analysis of different interpretability methods over a varying degree of feature correlation $\rho$. We present the performance of all methods on faithfulness (ROAR), monotonicity (ROAR), faithfulness, and infidelity (ordered from \textbf{left} to \textbf{right}) on the Gaussian Linear dataset for $\rho$ values ranging from [0, 1].}
    \label{fig:xaibench_correlation}
    %\vspace{-0.5cm}
\end{figure*}

%% file: Tables/interpretable_method_runtimes.tex
\begin{wraptable}{!tr}{0.5\textwidth}
%\begin{table}
\caption{Interpretable method inference times. All the methods are run on the GPU and the time is reported in seconds.}
\label{table:interpretable_method_runtimes}
\centering
\footnotesize
\begin{adjustbox}{width=0.4\textwidth}{
\begin{tabular}{@{}lccc@{}}
 \toprule
 \textbf{Method Name} & \textbf{Credit-g} & \textbf{Adult} & \textbf{Christine} \\
 \midrule
  IMN  & \textbf{0.01} & \textbf{0.02} & \textbf{0.02} \\
  TabNet  & 0.11 & 1.30 & 0.43 \\
  SHAP (TabResNet)  & 17.69 & 565.11 & 228.31 \\
  SHAP (CatBoost)  & 4.55 & 66.89 &  4317.61 \\
 \bottomrule 
\end{tabular}
}\end{adjustbox}
%\vspace{-0.5cm}
%\end{table}
\end{wraptable}

%% file: Tables/census_importances.tex
\begin{wraptable}{}{0.45\textwidth}
%\begin{table}
\caption{The feature rank importances for the Census dataset. A lower rank is associated with a higher feature importance.}
\label{table:census_importances}
\centering
\footnotesize
\begin{adjustbox}{width=0.45\textwidth}{
\begin{tabular}{@{}lccccc@{}}
 \toprule
 \textbf{Feature} & \textbf{SHAP} & \textbf{Decision Tree} & \textbf{TabNet} & \textbf{CatBoost} & \textbf{IMN}\\
 \midrule
  Age & \hl{2} & \hl{5} & \hl{2} & \hl{3} & \hl{3}  \\ 
  Capital Gain & 9 & \hl{4} & \hl{3} & \hl{1} & \hl{1} \\ 
  Capital Loss & 10 & 9 & 14 & \hl{4} & \hl{5} \\ 
  Demographic & \hl{1} & \hl{2} & 9 & 10 & 6 \\
  Education & \hl{5} & \hl{3} & \hl{5}  & 9 & 9 \\
  Education num. & \hl{4} & 12 & 6 & 6 & \hl{2} \\ 
  Hours per week & 6 & 7 & 7 & 7 & \hl{4} \\ 
  Race & 8 & 10 & 5 &  12 &  7 \\  
  Occupation & \hl{3} & 6  & 8 & 8 & 10 \\ 
  Relationship & 7 & \hl{1} & \hl{1} & \hl{2} & 8 \\
  
 \bottomrule 
\end{tabular}
}\end{adjustbox}
%\end{table}
\vspace{-0.3cm}
\end{wraptable}

%% file: Plots/importance_topk.tex
\begin{wrapfigure}{!hr}{0.45\textwidth}
    \centering
    \includegraphics[width=0.45\textwidth]
    {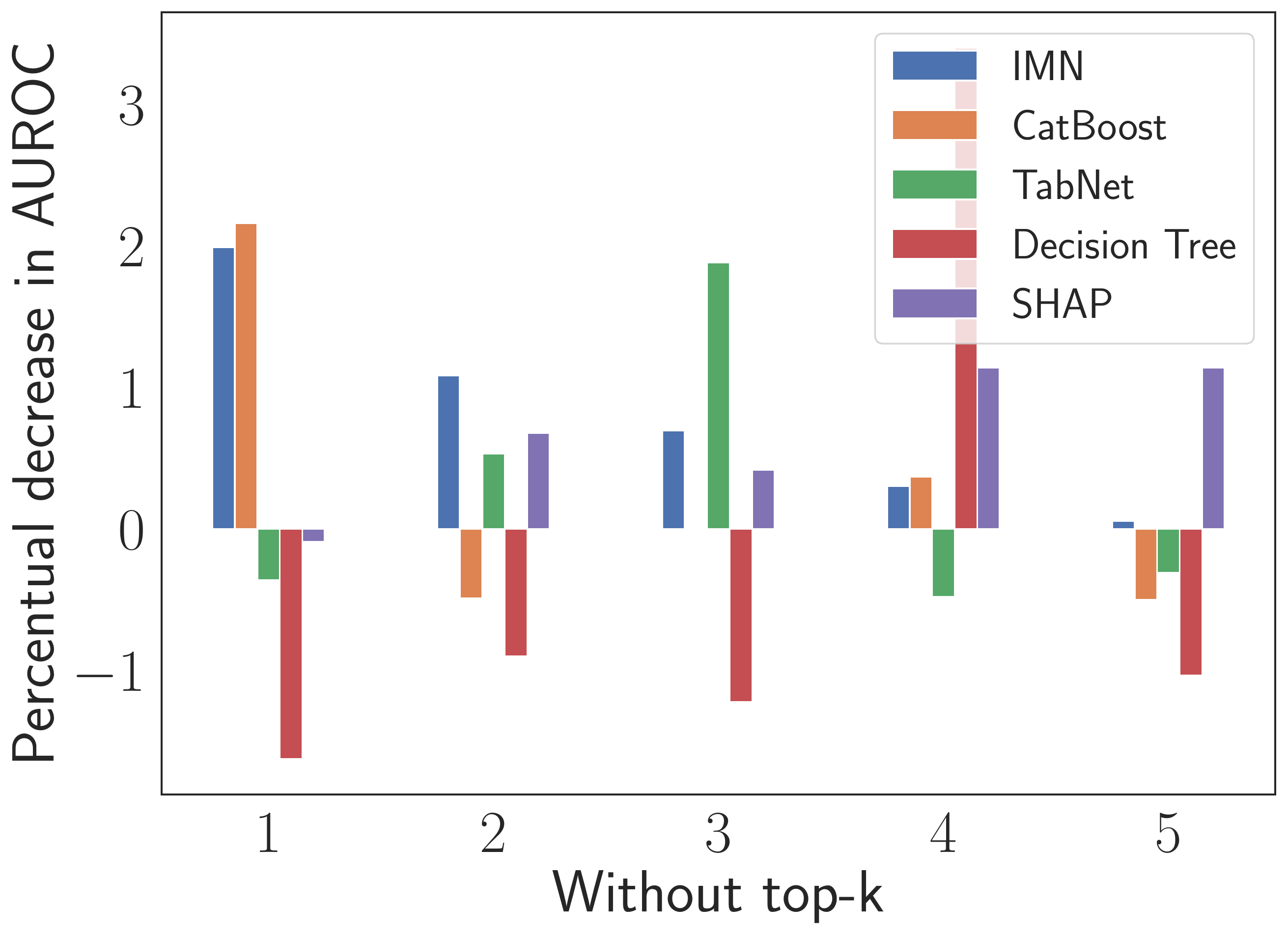}
    \caption[Top-k Feature Impact]{Investigating the decrease in AUROC when removing the $k$-th most important feature.}
    \label{fig:importance_topk}
    \vspace{-0.5cm}
\end{wrapfigure}

%% file: Plots/odor_feature_impacts.tex
\begin{wrapfigure}{!hr}{0.45\textwidth}
    \centering
    \includegraphics[width=0.45\textwidth]{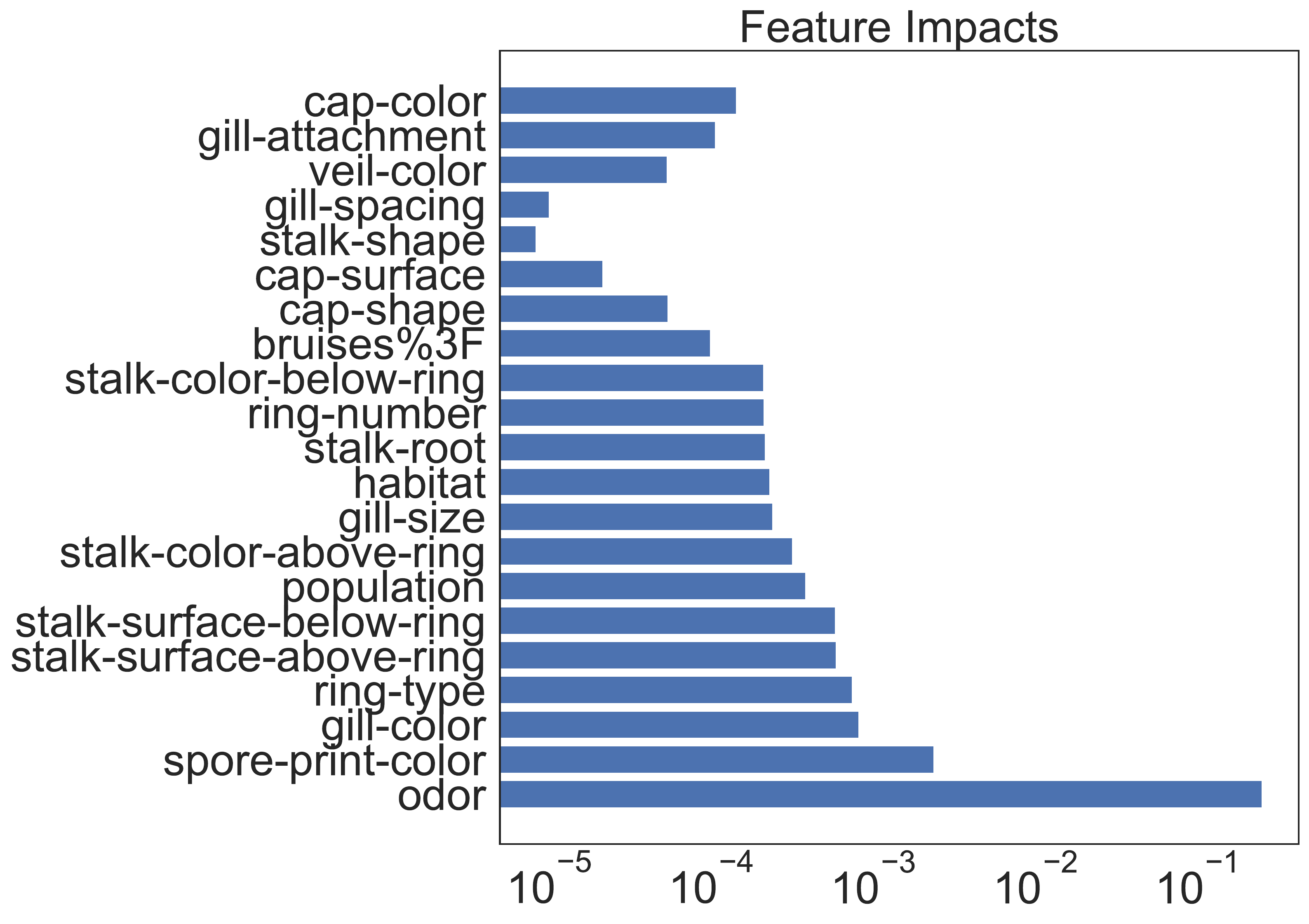}
    \caption{Feature impacts for the mushroom-edibility task.}
    \label{fig:odor_feature_impacts}
    \vspace{-1.1cm}
\end{wrapfigure}

%% file: Text/9-Conclusion.tex
\newpage

\section{Conclusion}

In this work, we propose explainable deep networks that are comparable in performance to their black-box counterparts but also as interpretable as state-of-the-art explanation techniques. With extensive experiments, we show that the explainable deep learning networks outperform traditional white-box models in terms of performance. Moreover, the experiments confirm that the explainable deep-learning architecture does not include a significant degradation in performance or an overhead on time compared to the plain black-box counterpart, achieving competitive results against state-of-the-art classifiers in tabular data. Our method matches competitive state-of-the-art explainability methods on a recent explainability benchmark in tabular data, offering explanations of predictions as a free lunch.

%% file: Text/10-Limitations.tex
\section{Limitations and Future Work}
\label{sec:limitations}

One potential limitation of our method is that although interpretable, the per-instance models are linear. A potential future work can focus on generating other types of non-linear interpretable models, such as decision trees. More concretely, the hypernetwork can generate the parameters of the decision splits and the decision value at each node, as well as the leaf weights. Another potential strategy is to generate local Support Vector Machines, by expressing the prediction for a data point as a function of the similarity of the informative neighbors. 

%% file: Text/11-Acknowledge.tex
\section*{Acknowledgements}

\textbf{JG}, \textbf{AK} and \textbf{SBA} would like to acknowledge the funding by the Deutsche Forschungsgemeinschaft (DFG, German Research Foundation) under grant number 417962828 and grant INST 39/963-1 FUGG (bwForCluster NEMO). In addition, \textbf{JG} and \textbf{AK} acknowledge the support of the BrainLinks-BrainTools center of excellence. Moreover, the authors acknowledge the support and HPC resources provided by the Erlangen National High Performance Computing Center (NHR@FAU) of the Friedrich-Alexander-Universität Erlangen-Nürnberg (FAU) under the NHR project v101be. NHR funding is provided by federal and Bavarian state authorities. NHR@FAU hardware is partially funded by the German Research Foundation (DFG) – 440719683.

%% file: Text/appendix-text.tex
\section{IMN can be extended to image classification backbones}
\label{appendix:image_application}

\begin{figure*}[!ht]
    \centering
    \includegraphics[width=0.9\textwidth]{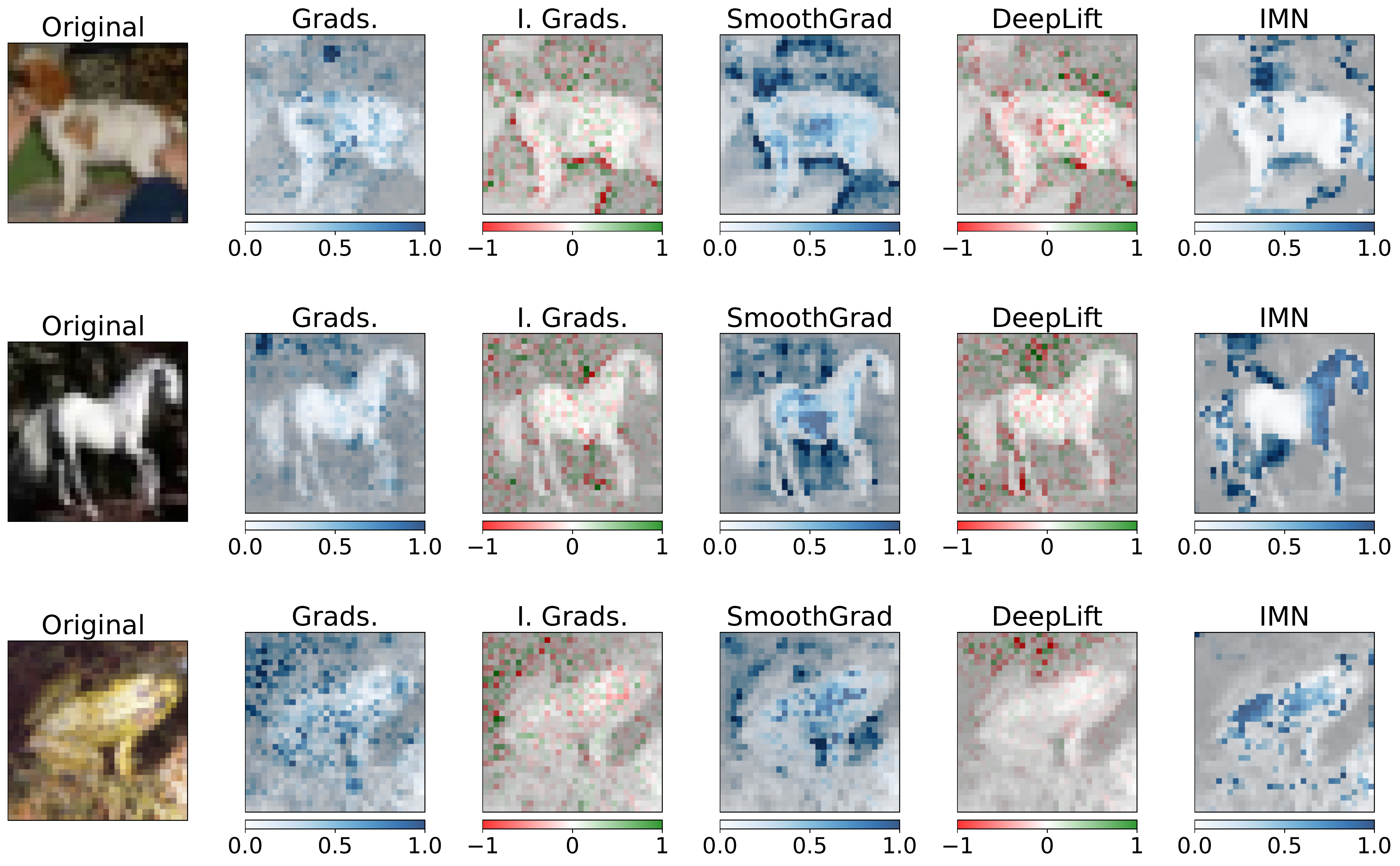}
    \caption{Comparison of IMN against explainability techniques for image classification.}
    \label{fig:Visualizing Interpretability for Computer Visions}
\end{figure*}

We use IMN to explain the predictions of ResNet50, a broadly used computer vision backbone. We take the pre-trained backbone $\phi(\cdot): \mathrm{R}^{H \times W \times K} \rightarrow \mathrm{R}^D $ from PyTorch and change the output layer to a fully-connected layer $w: \mathrm{R}^D \rightarrow \mathrm{R}^{H \times W \times K \times C}  $ that generates the weights for multiplying the input image $x \in \mathrm{R}^{H \times W \times K}$ with $K$ channels, and finally obtain the logits $z_c$ for the class $c$. In this experiment, we use $\lambda=10^{-3}$ as the L1 penalty strength. 

We fine-tuned the ImageNet pre-trained ResNet50 models, both for the explainable (IMN-ResNet) and the black-box (ResNet) variants for 400 epochs on the CIFAR-10 dataset with a learning rate of $10^{-4}$. To test whether the explainable variant is as accurate as the black-box model, we evaluate the validation accuracy after 5 independent training runs. IMN-ResNet achieves an accuracy of $87.49 \pm 1.73$ and the ResNet $88.76 \pm 1.50$, with the difference being statistically insignificant.  

We compare our method to the following image explainability baselines: Saliency Maps (Gradients)~\citep{simonyan2013deep}, DeepLift~\citep{ShrikumarGK17}, Integrated Gradients~\citep{ancona2017towards} with SmoothGrad. All of the baselines are available via the \textit{captum} library\footnote{\url{https://github.com/pytorch/captum}}. We compare the rival explainers to IMN-ResNet by visually interpreting the pixel-wise weights of selected images in Figure~\ref{fig:Visualizing Interpretability for Computer Visions}. The results confirm that IMN-ResNet generates higher weights for pixel regions that include descriptive parts of the object.

%% file: Text/appendix-plots.tex
\section{Plots}
\label{appendix:plots}

\input{Plots/cd_diagrams}

In Figure~\ref{fig:cd_diagram_default_hps}, we repeat the experiments from Hypotheses 1 and 2, however, without performing hyperparameter optimization. Moreover, we consider two additional baselines, DANet~\citep{chen2022danets} and HyperTab~\citep{wydmanski2023hypertab}. As observed, our findings are consistent and both hypotheses are validated even when default hyperparameters are used for all the methods considered.

\input{Plots/black_box_models_gain}

To further investigate the results on individual datasets, in Figure~\ref{fig:black_box_gain} we plot the distribution of the gains in performance of all methods over a single decision tree model (with default hyperparameters). The gain $G$ of a method $m$ run on a dataset $D$ for a single run is calculated as shown in Equation~\ref{eq:gain_calc}.

\begin{align}
    \label{eq:gain_calc}
    G\left(m, DTree, D\right) = &\frac{\text{AUROC}(m, D)}{\text{AUROC}(DTree, D)}
\end{align}

The results indicate that all methods except NAM achieve a comparable gain in performance across the AutoML benchmark datasets, while, the latter achieves a worse performance overall. We present detailed results in Appendix~\ref{appendix:tables}.  

Additionally, in Figure~\ref{fig:app_xaibench_correlation}, we present the performance of the different explainers for the different explainability metrics. We present results for the Gaussian Non-Linear Additive and Gaussian Piecewise Constant datasets over a varying presence of correlation $\rho$ between the features. The results show that our method achieves competitive results against Kernel Shap (K. SHAP) and LIME, the strongest baselines. 

\input{Plots/xaibench_correlation_appendix}

\begin{figure}[!ht]
  \centering
    \includegraphics[width=0.85\textwidth]{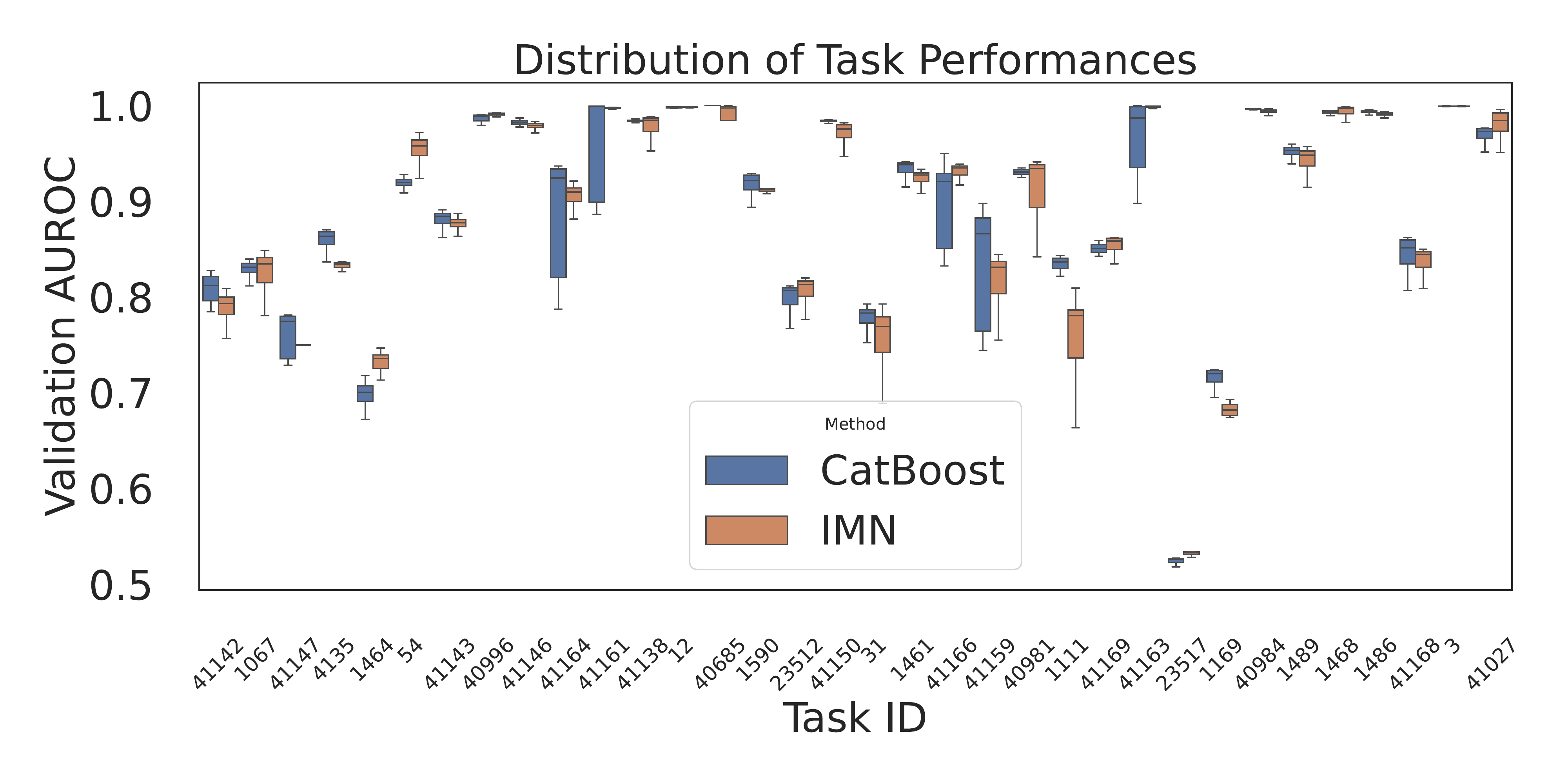}
    \caption{The distribution of the validation performance of the different hyperparameter configurations per task for CatBoost and IMN. }
    \label{app:hpo_performance_distribution}
\end{figure}

Lastly, to investigate how sensitive IMN is to the controlling hyperparameter configuration, we compare IMN and CatBoost (a method known for being robust to its hyperparameters in the community). Specifically, for every task, we plot the distribution of the performance of all hyperparameter configurations for every method. We present the results in Figure~\ref{app:hpo_performance_distribution}, where, as observed, IMN has a comparable sensitivity to CatBoost with regard to the controlling hyperparameter configuration. Moreover, in the majority of cases, the IMN validation performance does not vary significantly.

%% file: Plots/cd_diagrams.tex
\begin{figure}[!ht]
  \centering
    \includegraphics[width=0.3\textwidth]{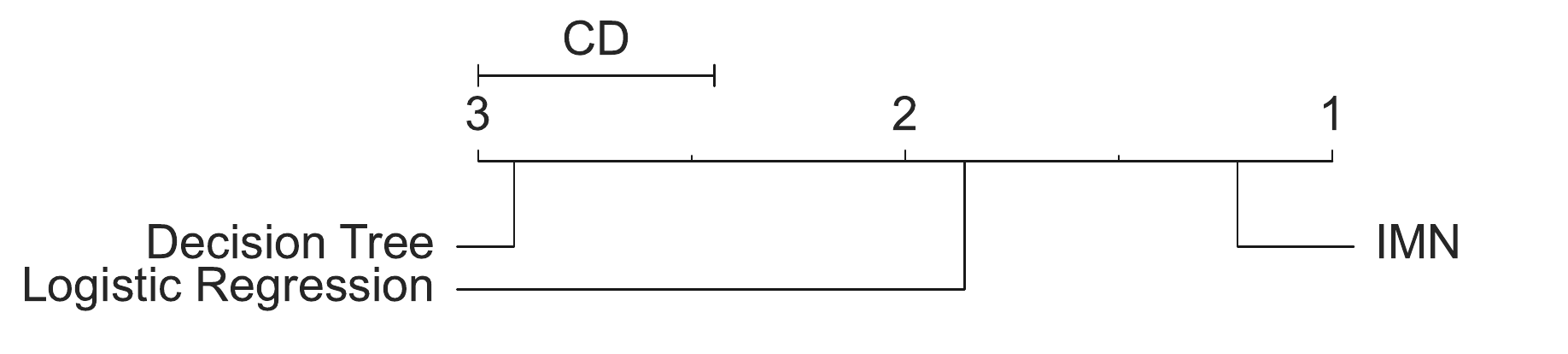}
    \includegraphics[width=0.3\textwidth]{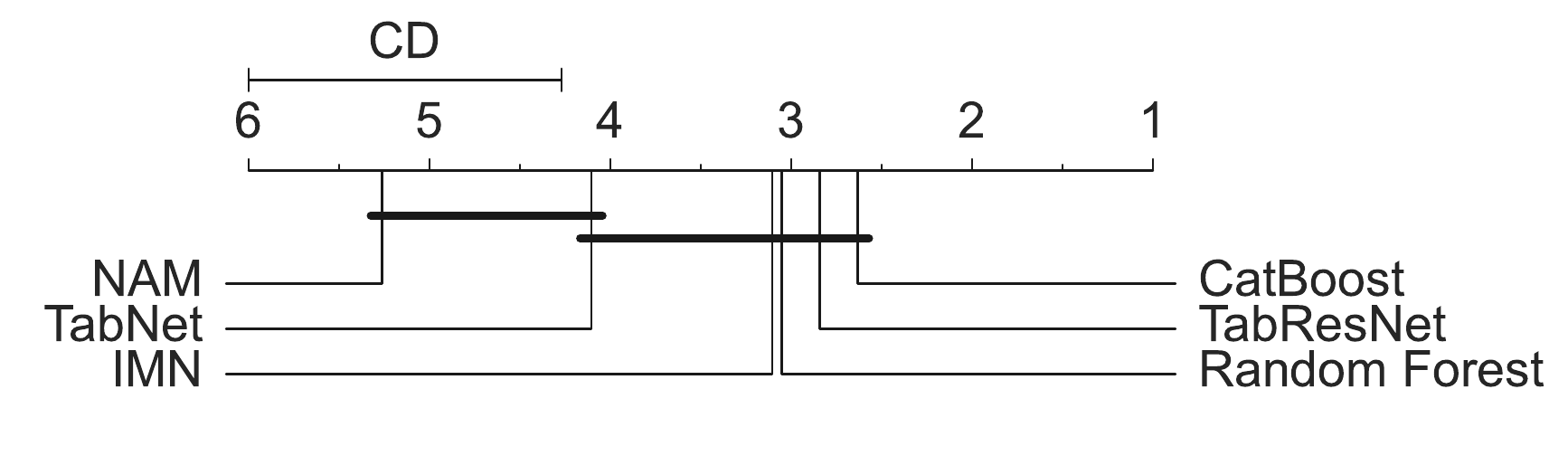}
    \includegraphics[width=0.3\textwidth]{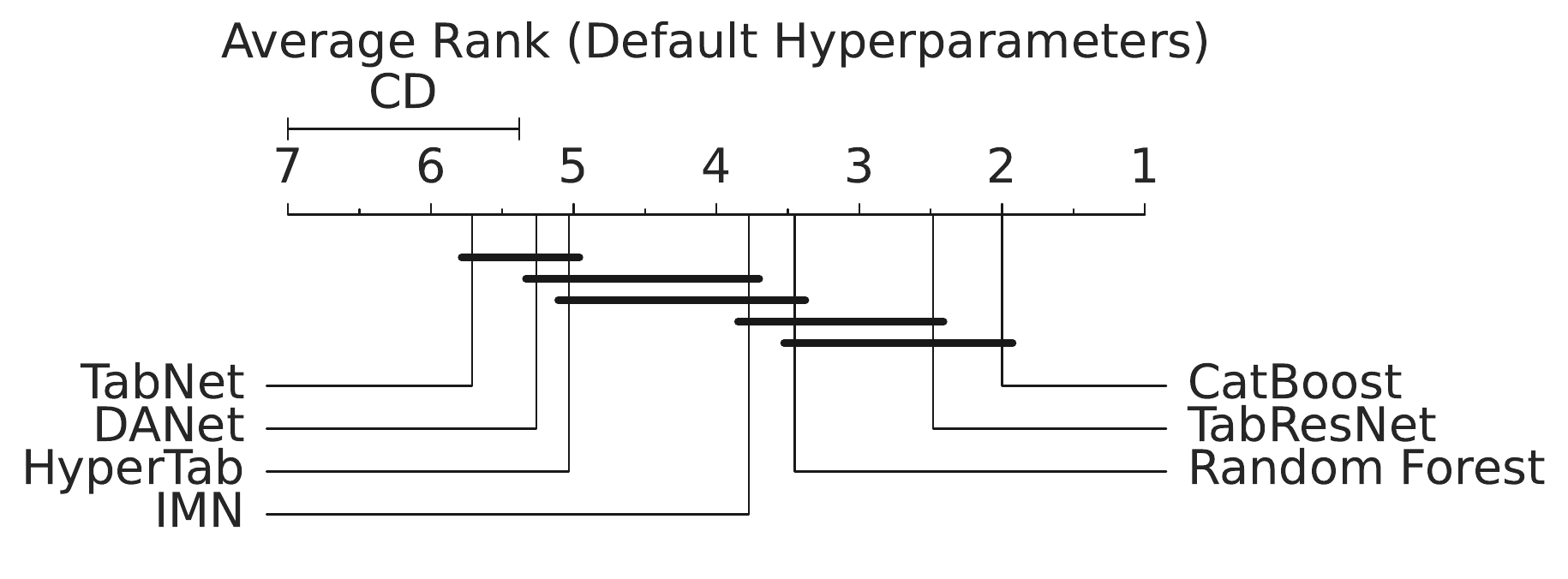}

  \caption{The critical difference diagrams that represent the average rank over all datasets based on the AUROC test performance for: \textbf{left)} The white-box methods and IMN, \textbf{middle)} The black-box methods and IMN for the binary classification datasets, \textbf{right)} The black-box methods and IMN for the entire benchmark of datasets.}
  \label{fig:cd_diagram_default_hps}
\end{figure}

%% file: Plots/black_box_models_gain.tex
\begin{wrapfigure}{!hr}{0.5\textwidth}
    \centering
    \includegraphics[width=0.5\textwidth]{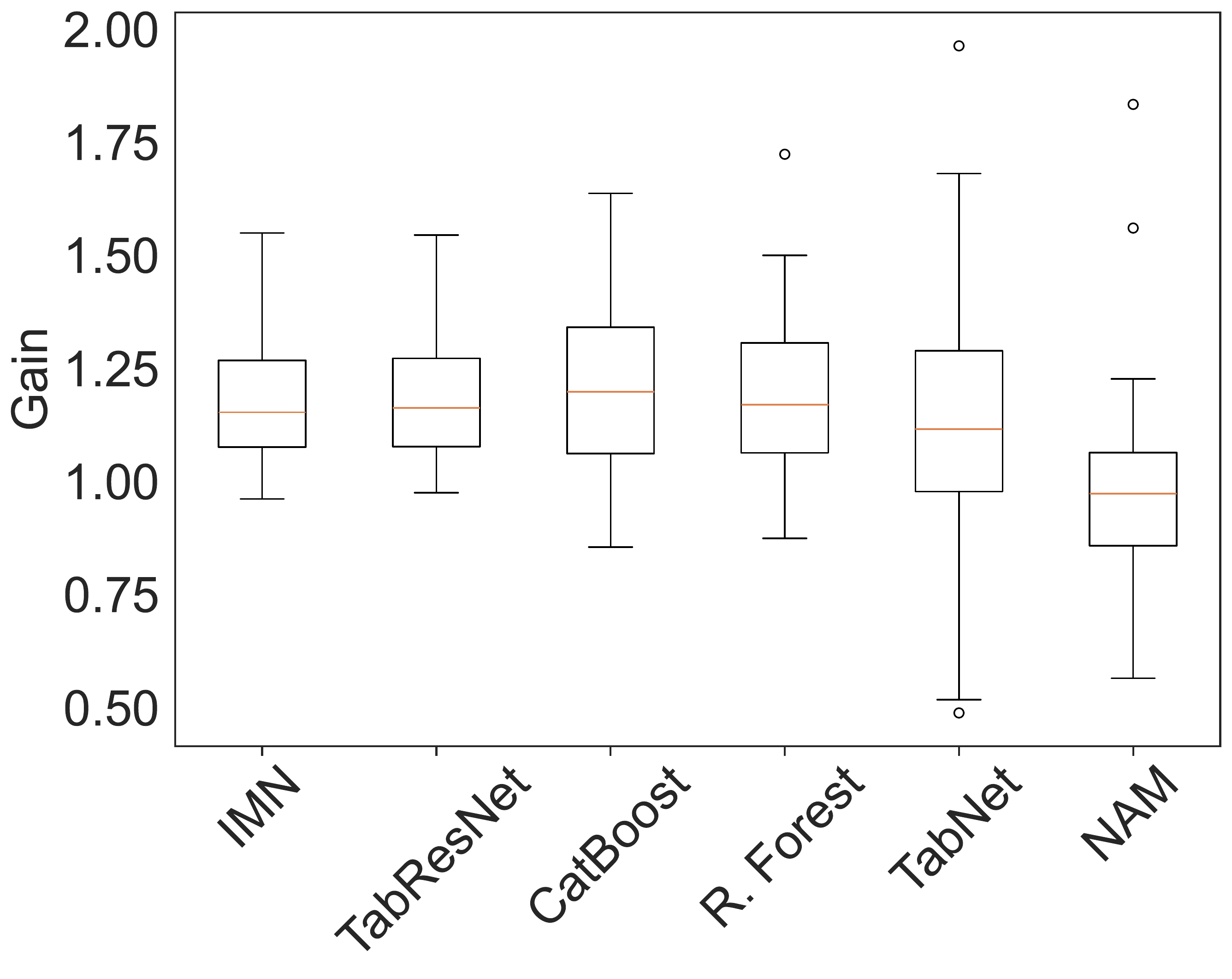}
    \caption[Black-box methods gain]{The gain distribution of the state-of-the-art models. The gain is calculated by dividing the test AUROC against the test AUROC of a decision tree.} 
    \label{fig:black_box_gain}
    \vspace{-0.5cm}
\end{wrapfigure}

%% file: Plots/xaibench_correlation_appendix.tex
\begin{figure*}[!ht]
    \centering 
\includegraphics[width=\textwidth]{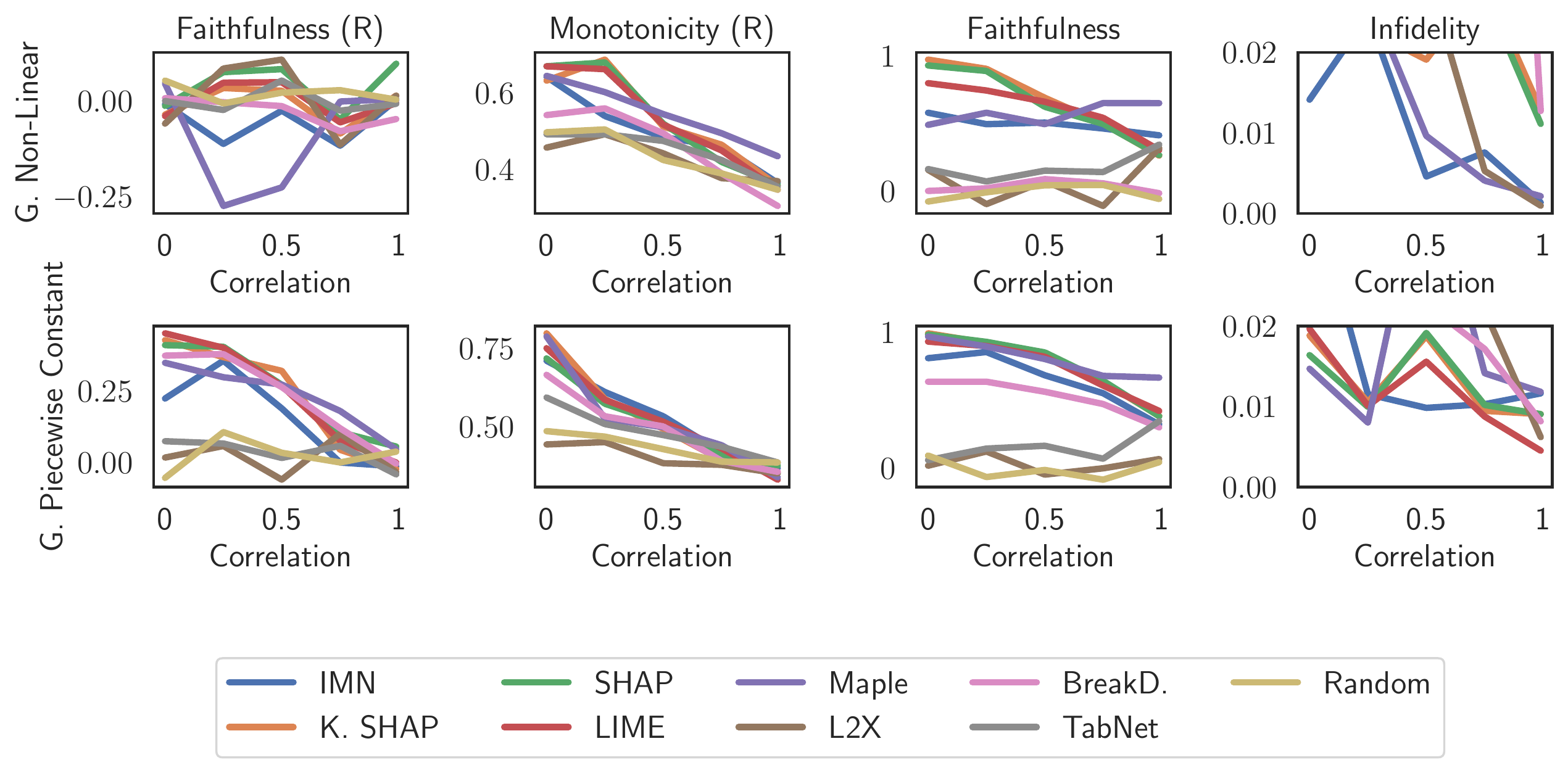}
    \caption[Interpretability analysis over correlation.]{Performance analysis of all explainable methods on faithfulness (ROAR), monotonicity (ROAR), faithfulness, and infidelity. The results are shown for the Gaussian Non-Linear Additive and Gaussian Piecewise datasets where, correlation ($\rho$) ranges from [0, 1].}
    \label{fig:app_xaibench_correlation}
    %\vspace{-0.5cm}
\end{figure*}

%% file: Text/appendix-tables.tex
\section{Tables}
\label{appendix:tables}

\bigskip

To describe the 35 datasets present in our accuracy-related experiments, we summarize the main descriptive statistics in Table~\ref{app:dataset_info}. The statistics show that our datasets are diverse, covering both binary and multi-class classification problems with imbalanced and balanced datasets that contain a diverse number of features and examples.

Additionally, we provide the per-dataset performances for the accuracy-related experiments of every method with the default configurations. Table~\ref{app:train_performances} summarizes the performances on the train split, where, as observed Random Forest and Decision Trees overfit the training data excessively compared to the other methods. Moreover, Table~\ref{app:test_performances} provides the performance of every method on the test split, where, IMN, TabResNet, and CatBoost achieve similar performances. We provide the same per-dataset performances of every method with the best-found hyperparameter configuration during HPO for the train split in Table~\ref{app:train_performances_tuned} and test split in Table~\ref{app:test_performances_tuned}.

Lastly, we provide the HPO search spaces of the different methods considered in our experiments in Table~\ref{app:imn_search_space}, \ref{app:logistic_regression_search_space}, \ref{app:decision_tree_search_space}, \ref{app:catboost_search_space},  \ref{app:random_forest_search_space}, \ref{app:tabnet_search_space}.

\input{Tables/dataset_information}
\input{Tables/all_train_performances}
\input{Tables/all_test_performances}
\input{Tables/all_train_performances_tuned}
\input{Tables/all_test_performances_tuned}

\bigskip
\clearpage
\newpage

\input{Tables/imn_search_space}
\input{Tables/logistic_regression_search_space}
\input{Tables/decision_tree_search_space}
\input{Tables/catboost_search_space}
\input{Tables/random_forest_search_space}
\input{Tables/tabnet_search_space}

%% file: Tables/dataset_information.tex
\begin{table}[!ht]
\centering
\caption{Statistics regarding the AutoML benchmark datasets.}
\label{app:dataset_info}
\footnotesize
\begin{adjustbox}{width=\textwidth}{
\begin{tabular}{llccccc}
\toprule
Dataset ID & Dataset Name & Number of Instances & Number of Features & Number of Classes & Majority Class Percentage & Minority Class Percentage \\
\midrule
3 & kr-vs-kp & 3196 & 37 & 2 & 52.222 & 47.778 \\
12 & mfeat-factors & 2000 & 217 & 10 & 10.000 & 10.000 \\
31 & credit-g & 1000 & 21 & 2 & 70.000 & 30.000 \\
54 & vehicle & 846 & 19 & 4 & 25.768 & 23.522 \\
1067 & kc1 & 2109 & 22 & 2 & 84.542 & 15.458 \\
1111 & KDDCup09 appetency & 50000 & 231 & 2 & 98.220 & 1.780 \\
1169 & airlines & 539383 & 8 & 2 & 55.456 & 44.544 \\
1461 & bank-marketing & 45211 & 17 & 2 & 88.302 & 11.698 \\
1464 & blood-transfusion-service-center & 748 & 5 & 2 & 76.203 & 23.797 \\
1468 & cnae-9 & 1080 & 857 & 9 & 11.111 & 11.111 \\
1486 & nomao & 34465 & 119 & 2 & 71.438 & 28.562 \\
1489 & phoneme & 5404 & 6 & 2 & 70.651 & 29.349 \\
1590 & adult & 48842 & 15 & 2 & 76.072 & 23.928 \\
4135 & Amazon employee access & 32769 & 10 & 2 & 94.211 & 5.789 \\
23512 & higgs & 98050 & 29 & 2 & 52.858 & 47.142 \\
23517 & numerai28.6 & 96320 & 22 & 2 & 50.517 & 49.483 \\
40685 & shuttle & 58000 & 10 & 7 & 78.597 & 0.017 \\
40981 & Australian & 690 & 15 & 2 & 55.507 & 44.493 \\
40984 & segment & 2310 & 20 & 7 & 14.286 & 14.286 \\
40996 & Fashion-MNIST & 70000 & 785 & 10 & 10.000 & 10.000 \\
41027 & jungle chess & 44819 & 7 & 3 & 51.456 & 9.672 \\
41138 & APSFailure & 76000 & 171 & 2 & 98.191 & 1.809 \\
41142 & christine & 5418 & 1637 & 2 & 50.000 & 50.000 \\
41143 & jasmine & 2984 & 145 & 2 & 50.000 & 50.000 \\
41146 & sylvine & 5124 & 21 & 2 & 50.000 & 50.000 \\
41147 & albert & 425240 & 79 & 2 & 50.000 & 50.000 \\
41150 & MiniBooNE & 130064 & 51 & 2 & 71.938 & 28.062 \\
41159 & guillermo & 20000 & 4297 & 2 & 59.985 & 40.015 \\
41161 & riccardo & 20000 & 4297 & 2 & 75.000 & 25.000 \\
41163 & dilbert & 10000 & 2001 & 5 & 20.490 & 19.130 \\
41164 & fabert & 8237 & 801 & 7 & 23.394 & 6.094 \\
41165 & robert & 10000 & 7201 & 10 & 10.430 & 9.580 \\
41166 & volkert & 58310 & 181 & 10 & 21.962 & 2.334 \\
41168 & jannis & 83733 & 55 & 4 & 46.006 & 2.015 \\
41169 & helena & 65196 & 28 & 100 & 6.143 & 0.170 \\
\bottomrule
\end{tabular}}
\end{adjustbox}
\end{table}

%% file: Tables/all_train_performances.tex
\begin{table}[!ht]
\centering
\caption{The per-dataset train AUROC performance for all methods in the accuracy experiments with default hyperparameter configurations. The train performance is the mean value from 10 runs with different seeds. A dashed line '-' represents a failure of running on that particular dataset.}
\label{app:train_performances}
\footnotesize
\begin{adjustbox}{width=\textwidth}{
\begin{tabular}{lcccccccc}
\toprule
Dataset ID & Decision Tree & Logistic Regression & Random Forest & TabNet & TabResNet & CatBoost & IMN \\
\midrule
3 & \textbf{1.000} & 0.990 & \textbf{1.000} & 0.980 & \textbf{1.000} & \textbf{1.000} & \textbf{1.000} \\
12 & \textbf{1.000} & \textbf{1.000} & \textbf{1.000} & \textbf{1.000} & \textbf{1.000} & \textbf{1.000} & \textbf{1.000} \\
31 & \textbf{1.000} & 0.795 & \textbf{1.000} & 0.514 & \textbf{1.000} & 0.963 & \textbf{1.000} \\
54 & \textbf{1.000} & 0.955 & \textbf{1.000} & 0.492 & \textbf{1.000} & \textbf{1.000} & \textbf{1.000} \\
1067 & \textbf{0.998} & 0.818 & 0.997 & 0.825 & 0.928 & 0.971 & 0.920 \\
1111 & \textbf{1.000} & 0.822 & \textbf{1.000} & - & 0.966 & 0.899 & 0.895 \\
1169 & \textbf{0.994} & 0.680 & \textbf{0.994} & 0.705 & 0.697 & 0.733 & 0.697 \\
1461 & \textbf{1.000} & 0.908 & \textbf{1.000} & 0.947 & 0.945 & 0.948 & 0.942 \\
1464 & 0.983 & 0.757 & 0.978 & 0.490 & 0.830 & \textbf{0.934} & 0.834 \\
1468 & \textbf{1.000} & \textbf{1.000} & \textbf{1.000} & 0.493 & \textbf{1.000} & \textbf{1.000} & \textbf{1.000} \\
1486 & \textbf{1.000} & 0.988 & \textbf{1.000} & 0.995 & 0.999 & 0.997 & 0.998 \\
1489 & \textbf{1.000} & 0.813 & \textbf{1.000} & 0.947 & 0.974 & 0.982 & 0.977 \\
1590 & \textbf{1.000} & 0.903 & \textbf{1.000} & 0.920 & 0.920 & 0.935 & 0.920 \\
1596 & \textbf{1.000} & 0.951 & \textbf{1.000} & 0.949 & 0.992 & 0.997 & 0.995 \\
4135 & \textbf{1.000} & 0.839 & 0.998 & - & 0.890 & 0.981 & 0.877 \\
23512 & \textbf{1.000} & 0.683 & \textbf{1.000} & 0.820 & 0.863 & 0.831 & 0.853 \\
23517 & \textbf{1.000} & 0.533 & \textbf{1.000} & 0.529 & 0.587 & 0.703 & 0.546 \\
40685 & \textbf{1.000} & 0.999 & \textbf{1.000} & 0.988 & 0.999 & - & 0.994 \\
40981 & \textbf{1.000} & 0.932 & \textbf{1.000} & 0.472 & \textbf{1.000} & 0.996 & \textbf{1.000} \\
40984 & \textbf{1.000} & 0.983 & \textbf{1.000} & 0.990 & 0.999 & \textbf{1.000} & 0.999 \\
40996 & \textbf{1.000} & 0.989 & \textbf{1.000} & 0.997 & \textbf{1.000} & 0.999 & \textbf{1.000} \\
41027 & \textbf{1.000} & 0.799 & \textbf{1.000} & 0.980 & 0.982 & 0.989 & 0.982 \\
41138 & \textbf{1.000} & 0.992 & \textbf{1.000} & 0.999 & 0.999 & \textbf{1.000} & 0.996 \\
41142 & \textbf{1.000} & 0.942 & \textbf{1.000} & 0.951 & \textbf{1.000} & 0.999 & \textbf{1.000} \\
41143 & \textbf{1.000} & 0.868 & \textbf{1.000} & 0.874 & \textbf{1.000} & 0.992 & \textbf{1.000} \\
41146 & \textbf{1.000} & 0.967 & \textbf{1.000} & 0.989 & \textbf{1.000} & \textbf{1.000} & \textbf{1.000} \\
41147 & \textbf{1.000} & 0.746 & \textbf{1.000} & - & 0.769 & 0.827 & 0.763 \\
41150 & \textbf{1.000} & 0.938 & \textbf{1.000} & 0.896 & 0.985 & 0.988 & 0.985 \\
41159 & \textbf{1.000} & 0.826 & \textbf{1.000} & 0.840 & \textbf{1.000} & 0.977 & \textbf{1.000} \\
41161 & \textbf{1.000} & \textbf{1.000} & \textbf{1.000} & 0.999 & \textbf{1.000} & \textbf{1.000} & \textbf{1.000} \\
41163 & \textbf{1.000} & \textbf{1.000} & \textbf{1.000} & \textbf{1.000} & \textbf{1.000} & \textbf{1.000} & \textbf{1.000} \\
41164 & \textbf{1.000} & 0.994 & \textbf{1.000} & 0.968 & \textbf{1.000} & 0.983 & \textbf{1.000} \\
41165 & \textbf{1.000} & \textbf{1.000} & \textbf{1.000} & 0.876 & \textbf{1.000} & \textbf{1.000} & \textbf{1.000} \\
41166 & \textbf{1.000} & 0.889 & \textbf{1.000} & 0.943 & 0.978 & 0.992 & 0.995 \\
41168 & \textbf{1.000} & 0.804 & \textbf{1.000} & 0.911 & 0.915 & 0.971 & 0.921 \\
41169 & \textbf{1.000} & 0.854 & \textbf{1.000} & 0.867 & 0.938 & 0.998 & 0.986 \\
\bottomrule
\end{tabular}}
\end{adjustbox}
\end{table}

%% file: Tables/all_test_performances.tex
\begin{table}[!ht]
\centering
\caption{The per-dataset test AUROC performance for all methods in the accuracy experiments with default hyperparameter configurations. The test performance is the mean value from 10 runs with different seeds. A dashed line '-' represents a failure of running on that particular dataset.}
\label{app:test_performances}
\footnotesize
\begin{adjustbox}{width=\textwidth}{
\begin{tabular}{lcccccccc}
\toprule
Dataset ID & Decision Tree & Logistic Regression & NAM & Random Forest & TabNet & TabResNet & CatBoost & IMN \\
\midrule
3 & 0.987 & 0.990 & 0.977 & 0.998 & 0.983 & \textbf{0.999} & \textbf{0.999} & \textbf{0.999} \\
12 & 0.938 & \textbf{0.999} & - & 0.998 & 0.995 & \textbf{0.999} & \textbf{0.999} & \textbf{0.999} \\
31 & 0.643 & 0.775 & 0.717 & \textbf{0.795} & 0.511 & 0.756 & 0.790 & 0.751 \\
54 & 0.804 & 0.938 & - & 0.927 & 0.501 & \textbf{0.968} & 0.934 & 0.957 \\
1067 & 0.620 & 0.802 & 0.659 & 0.801 & 0.789 & \textbf{0.808} & 0.800 & 0.805 \\
1111 & 0.535 & 0.816 & 0.544 & 0.793 & - & 0.778 & \textbf{0.843} & 0.816 \\
1169 & 0.592 & 0.679 & 0.588 & 0.692 & \textbf{0.699} & 0.695 & \textbf{0.718} & 0.695 \\
1461 & 0.703 & 0.908 & 0.827 & 0.930 & 0.926 & 0.931 & \textbf{0.937} & 0.930 \\
1464 & 0.599 & 0.749 & 0.738 & 0.666 & 0.516 & 0.740 & 0.709 & \textbf{0.742} \\
1468 & 0.926 & \textbf{0.996} & - & 0.995 & 0.495 & 0.995 & \textbf{0.996} & 0.994 \\
1486 & 0.935 & 0.987 & 0.934 & 0.993 & 0.991 & \textbf{0.994} & \textbf{0.994} & 0.993 \\
1489 & 0.842 & 0.805 & 0.806 & \textbf{0.962} & 0.928 & 0.949 & 0.948 & \textbf{0.950} \\
1590 & 0.752 & 0.903 & 0.874 & 0.917 & 0.908 & 0.915 & \textbf{0.930} & 0.915 \\
1596 & 0.942 & 0.951 & - & \textbf{0.997} & 0.949 & 0.991 & 0.996 & 0.994 \\
4135 & 0.639 & 0.853 & 0.838 & 0.846 & - & 0.855 & \textbf{0.883} & 0.858 \\
23512 & 0.626 & 0.683 & 0.583 & 0.794 & 0.803 & \textbf{0.825} & 0.804 & 0.823 \\
23517 & 0.501 & 0.530 & 0.505 & 0.515 & 0.522 & 0.529 & 0.526 & \textbf{0.530} \\
40685 & 0.967 & 0.994 & - & \textbf{1.000} & 0.986 & 0.995 & - & 0.993 \\
40981 & 0.817 & 0.930 & 0.918 & \textbf{0.945} & 0.463 & 0.919 & 0.935 & 0.908 \\
40984 & 0.946 & 0.980 & - & \textbf{0.995} & 0.985 & 0.994 & \textbf{0.995} & 0.994 \\
40996 & 0.886 & 0.984 & - & 0.991 & 0.989 & \textbf{0.994} & 0.993 & 0.992 \\
41027 & 0.792 & 0.797 & - & 0.931 & 0.976 & \textbf{0.979} & 0.974 & 0.978 \\
41138 & 0.861 & 0.974 & 0.558 & 0.989 & 0.970 & 0.972 & \textbf{0.992} & 0.980 \\
41142 & 0.626 & 0.742 & 0.724 & 0.796 & 0.713 & 0.782 & \textbf{0.822} & 0.775 \\
41143 & 0.749 & 0.850 & 0.831 & 0.880 & 0.823 & 0.860 & \textbf{0.870} & 0.865 \\
41146 & 0.910 & 0.966 & - & 0.983 & 0.974 & 0.982 & \textbf{0.988} & 0.981 \\
41147 & 0.606 & 0.748 & 0.675 & 0.762 & - & 0.765 & \textbf{0.779} & 0.762 \\
41150 & 0.867 & 0.938 & 0.912 & 0.981 & 0.896 & \textbf{0.984} & \textbf{0.984} & \textbf{0.984} \\
41159 & 0.730 & 0.712 & 0.618 & \textbf{0.892} & 0.754 & 0.871 & \textbf{0.897} & 0.841 \\
41161 & 0.857 & 0.995 & 0.972 & 0.999 & 0.997 & 0.998 & \textbf{1.000} & 0.998 \\
41163 & 0.873 & 0.994 & - & 0.999 & 0.998 & \textbf{1.000} & \textbf{1.000} & \textbf{1.000} \\
41164 & 0.786 & 0.898 & - & 0.925 & 0.888 & 0.913 & \textbf{0.935} & 0.902 \\
41165 & 0.579 & 0.748 & - & 0.835 & 0.788 & 0.838 & \textbf{0.895} & 0.817 \\
41166 & 0.699 & 0.882 & - & 0.927 & 0.918 & \textbf{0.952} & 0.949 & 0.943 \\
41168 & 0.633 & 0.798 & - & 0.831 & 0.813 & \textbf{0.868} & 0.862 & 0.856 \\
41169 & 0.554 & 0.841 & - & 0.800 & 0.842 & \textbf{0.883} & 0.866 & 0.865 \\
\bottomrule
\end{tabular}}
\end{adjustbox}
\end{table}

%% file: Tables/all_train_performances_tuned.tex
\begin{table}[!ht]
\centering
\caption{The per-dataset train AUROC performance for all methods in the accuracy experiments parametrized with the best hyperparameter configuration found during HPO. A dashed line '-' represents a failure to run on that particular dataset.}
\label{app:train_performances_tuned}
\footnotesize
\begin{adjustbox}{width=\textwidth}{
\begin{tabular}{lcccccccc}
\toprule
Dataset ID & Decision Tree & Logistic Regression & Random Forest & TabNet & TabResNet & CatBoost & IMN \\ \midrule
3 & 0.999 & 0.995 & 0.997 & - & 1.000 & \textbf{1.000} & 1.000 \\
12 & 0.986 & \textbf{1.000} & 0.999 & 1.000 & \textbf{1.000} & \textbf{1.000} & \textbf{1.000} \\
31 & 0.820 & 0.785 & 0.945 & \textbf{0.982} & 0.806 & 0.888 & 0.813 \\
54 & 0.930 & 0.960 & 0.971 & 0.988 & 0.990 & \textbf{1.000} & 0.968 \\
1067 & 0.831 & 0.807 & 0.819 & \textbf{0.891} & 0.813 & 0.875 & 0.810 \\
1111 & 0.836 & 0.829 & \textbf{0.885} & - & 0.825 & 0.842 & 0.844 \\
1169 & 0.682 & 0.680 & 0.690 & 0.714 & 0.696 & \textbf{0.754} & 0.683 \\
1461 & 0.900 & 0.907 & 0.922 & 0.946 & 0.947 & \textbf{0.954} & 0.944 \\
1464 & 0.796 & 0.765 & 0.867 & 0.835 & 0.811 & \textbf{0.947} & 0.763 \\
1468 & 0.972 & 1.000 & 0.996 & 1.000 & \textbf{1.000} & 1.000 & \textbf{1.000} \\
1486 & 0.981 & 0.988 & 0.987 & 0.997 & 0.998 & \textbf{0.999} & 0.997 \\
1489 & 0.908 & 0.815 & 0.938 & 0.993 & 0.995 & \textbf{1.000} & 0.998 \\
1590 & 0.904 & 0.903 & 0.911 & - & 0.920 & \textbf{0.943} & 0.926 \\
1596 & 0.931 & 0.951 & 0.946 & 0.952 & - & \textbf{0.998} & - \\
4135 & 0.833 & 0.826 & 0.865 & - & 0.843 & \textbf{0.995} & 0.850 \\
23512 & 0.737 & 0.684 & 0.766 & 0.831 & 0.869 & \textbf{0.900} & 0.859 \\
23517 & 0.529 & 0.532 & 0.562 & 0.523 & 0.531 & \textbf{0.587} & 0.535 \\
40685 & 1.000 & 0.999 & 1.000 & 1.000 & 1.000 & \textbf{1.000} & 0.996 \\
40981 & 0.931 & 0.932 & \textbf{0.978} & - & 0.949 & 0.943 & 0.947 \\
40984 & 0.989 & 0.988 & 0.996 & 0.999 & 0.999 & \textbf{1.000} & 0.999 \\
40996 & 0.956 & 0.988 & 0.976 & 0.998 & \textbf{1.000} & 0.994 & - \\
41027 & 0.873 & 0.801 & 0.905 & \textbf{0.999} & 0.996 & 0.989 & 0.996 \\
41138 & 0.979 & 0.989 & 0.989 & 0.996 & 0.992 & \textbf{1.000} & 0.991 \\
41142 & 0.786 & 0.874 & 0.860 & - & 0.976 & 0.952 & \textbf{1.000} \\
41143 & 0.849 & 0.872 & 0.934 & 0.877 & 0.977 & \textbf{0.997} & 0.941 \\
41146 & 0.966 & 0.968 & 0.986 & \textbf{1.000} & 1.000 & 0.999 & 0.994 \\
41147 & 0.727 & 0.745 & 0.746 & - & 0.768 & \textbf{0.865} & 0.750 \\
41150 & 0.946 & 0.956 & 0.961 & 0.968 & 0.989 & \textbf{1.000} & 0.989 \\
41159 & 0.775 & 0.823 & 0.878 & 0.552 & \textbf{1.000} & \textbf{1.000} & 1.000 \\
41161 & 0.860 & 0.999 & 0.961 & 0.999 & 1.000 & \textbf{1.000} & 1.000 \\
41163 & 0.901 & 1.000 & 0.976 & 1.000 & \textbf{1.000} & \textbf{1.000} & \textbf{1.000} \\
41164 & 0.707 & 0.975 & 0.901 & 0.998 & 0.982 & 0.999 & \textbf{0.999} \\
41165 & 0.750 & 0.936 & 0.840 & 0.960 & \textbf{0.995} & 0.988 & - \\
41166 & 0.822 & 0.892 & 0.867 & 0.986 & 0.984 & \textbf{1.000} & 0.999 \\
41167 & 0.941 & 0.997 & - & \textbf{1.000} & - & - & - \\
41168 & 0.787 & 0.804 & 0.823 & 0.908 & 0.900 & \textbf{0.944} & 0.908 \\
41169 & 0.791 & 0.853 & 0.846 & 0.879 & 0.926 & \textbf{0.976} & 0.961 \\
\bottomrule
\end{tabular}}
\end{adjustbox}
\end{table}

%% file: Tables/all_test_performances_tuned.tex
\begin{table}[!ht]
\centering
\caption{The per-dataset test AUROC performance for all methods in the accuracy experiments parametrized with the best hyperparameter configuration found during HPO. A dashed line '-' represents a failure to run on that particular dataset.}
\label{app:test_performances_tuned}
\footnotesize
\begin{adjustbox}{width=\textwidth}{
\begin{tabular}{lcccccccc}
\toprule
Dataset ID & Decision Tree & Logistic Regression & NAM & Random Forest & TabNet & TabResNet & CatBoost & IMN \\
\midrule
3 & 0.999 & 0.997 & 0.977 & 0.998 & - & 1.000 & 1.000 & \textbf{1.000} \\
12 & 0.960 & 0.998 & - & 0.998 & 0.998 & 0.998 & 0.998 & \textbf{1.000} \\
31 & 0.765 & 0.852 & 0.717 & 0.826 & 0.741 & 0.844 & 0.826 & \textbf{0.863} \\
54 & 0.844 & 0.959 & - & 0.924 & 0.955 & \textbf{0.966} & 0.935 & 0.959 \\
1067 & 0.777 & 0.804 & 0.659 & 0.813 & 0.784 & 0.795 & \textbf{0.834} & 0.805 \\
1111 & 0.805 & 0.814 & 0.544 & 0.829 & - & 0.801 & \textbf{0.839} & 0.806 \\
1169 & 0.679 & 0.678 & 0.588 & 0.686 & 0.702 & 0.693 & \textbf{0.724} & 0.680 \\
1461 & 0.903 & 0.907 & 0.827 & 0.919 & 0.917 & 0.932 & \textbf{0.939} & 0.933 \\
1464 & 0.647 & 0.729 & \textbf{0.738} & 0.679 & 0.668 & 0.697 & 0.678 & 0.710 \\
1468 & 0.954 & 0.999 & - & 0.988 & 0.987 & 0.999 & 0.994 & \textbf{0.999} \\
1486 & 0.977 & 0.986 & 0.934 & 0.985 & 0.989 & 0.994 & \textbf{0.996} & 0.992 \\
1489 & 0.890 & 0.802 & 0.806 & 0.920 & 0.945 & \textbf{0.958} & 0.955 & 0.956 \\
1590 & 0.901 & 0.901 & 0.874 & 0.908 & - & 0.913 & \textbf{0.930} & 0.913 \\
1596 & 0.931 & 0.951 & - & 0.946 & 0.952 & - & \textbf{0.997} & - \\
4135 & 0.845 & 0.854 & 0.838 & 0.870 & - & 0.861 & \textbf{0.909} & 0.865 \\
23512 & 0.725 & 0.681 & 0.583 & 0.754 & 0.803 & \textbf{0.817} & 0.808 & 0.817 \\
23517 & 0.522 & 0.530 & 0.505 & 0.529 & 0.525 & 0.527 & \textbf{0.532} & 0.531 \\
40685 & 0.908 & 0.999 & - & 1.000 & 1.000 & 0.999 & \textbf{1.000} & 1.000 \\
40981 & 0.905 & 0.916 & 0.918 & \textbf{0.933} & - & 0.916 & 0.920 & 0.909 \\
40984 & 0.981 & 0.989 & - & 0.991 & 0.995 & 0.994 & \textbf{0.996} & 0.993 \\
40996 & 0.954 & 0.985 & - & 0.976 & 0.989 & \textbf{0.994} & 0.991 & - \\
41027 & 0.870 & 0.800 & - & 0.900 & \textbf{0.999} & 0.993 & 0.980 & 0.991 \\
41138 & 0.974 & 0.986 & 0.558 & 0.985 & \textbf{0.987} & 0.986 & 0.986 & 0.986 \\
41142 & 0.744 & 0.811 & 0.724 & 0.794 & - & 0.814 & \textbf{0.819} & 0.808 \\
41143 & 0.843 & 0.839 & 0.831 & 0.871 & 0.846 & 0.860 & \textbf{0.871} & 0.856 \\
41146 & 0.966 & 0.962 & - & 0.975 & \textbf{0.991} & 0.984 & 0.987 & 0.984 \\
41147 & 0.728 & 0.748 & 0.675 & 0.748 & - & 0.767 & \textbf{0.786} & 0.754 \\
41150 & 0.942 & 0.954 & 0.912 & 0.958 & 0.967 & 0.985 & \textbf{0.986} & 0.984 \\
41159 & 0.753 & 0.725 & 0.618 & 0.847 & 0.543 & 0.871 & \textbf{0.912} & 0.868 \\
41161 & 0.849 & 0.997 & 0.972 & 0.950 & 0.996 & 0.999 & \textbf{0.999} & 0.998 \\
41163 & 0.885 & 0.996 & - & 0.969 & 0.997 & 1.000 & \textbf{1.000} & 1.000 \\
41164 & 0.697 & 0.914 & - & 0.877 & 0.869 & 0.919 & \textbf{0.937} & 0.919 \\
41165 & 0.706 & 0.822 & - & 0.804 & 0.777 & 0.870 & \textbf{0.875} & - \\
41166 & 0.811 & 0.886 & - & 0.862 & 0.929 & 0.954 & \textbf{0.956} & 0.947 \\
41168 & 0.768 & 0.797 & - & 0.809 & 0.808 & 0.866 & \textbf{0.871} & 0.856 \\
41169 & 0.762 & 0.845 & - & 0.823 & 0.839 & \textbf{0.884} & 0.869 & 0.873 \\
\bottomrule
\end{tabular}}
\end{adjustbox}
\end{table}

%% file: Tables/imn_search_space.tex
\begin{table}[ht]
\caption{\label{app:imn_search_space}The hyperparameter search space for IMN. TabResNet has the same search space without weight normalization.}
\begin{center}
\footnotesize
\resizebox{0.6\linewidth}{!}{
%\begin{adjustbox}{width=\textwidth}{
\begin{tabular}{@{}llcc@{}}
 \toprule
 \textbf{Hyperparameter} & \textbf{Type} & \textbf{Range} & \textbf{Log scale} \\
 \midrule
  $nr\_epochs$ & Integer & $[10, 500]$ &  -\\ \midrule
 $learning\_rate$ & Continuous & [1e-5, 1e-1] & \checkmark  \\ \midrule
 $batch\_size$ & Categorical & $\{32, 64, 128, 256, 512\}$ & -  \\ \midrule
 $weight\_decay$ & Continuous & $[1e-5, 1e-1]$ & \checkmark \\ \midrule
 $weight\_norm$ & Continuous & $[1e-5, 1e-1]$ & \checkmark \\ \midrule
 $dropout\_rate$ & Continuous & $[0, 0.5]$ & - \\ \bottomrule 
\end{tabular}}
%}\end{adjustbox}
\end{center}

\end{table}

%% file: Tables/logistic_regression_search_space.tex
\begin{table}[ht]
\caption{\label{app:logistic_regression_search_space} The hyperparameter search space for logistic regression.}
\begin{center}
\footnotesize
\resizebox{0.6\linewidth}{!}{
%\begin{adjustbox}{width=\textwidth}{
\begin{tabular}{@{}llcc@{}}
 \toprule
 \textbf{Hyperparameter} & \textbf{Type} & \textbf{Range} & \textbf{Log scale} \\
 \midrule
  $C$ & Continuous & $[1e-5, 5]$ &  -\\ \midrule
 $penalty$ & Categorical & $\{l2, none\}$ & -  \\ \midrule
 $max\_iterations$ & Integer & $[50, 500]$ & -  \\ \midrule
 $fit\_intercept$ & Categorical & $\{True, False\}$ & -  \\ \bottomrule 
\end{tabular}}
%}\end{adjustbox}
\end{center}

\end{table}

%% file: Tables/decision_tree_search_space.tex
\begin{table}[ht]
\caption{\label{app:decision_tree_search_space} The hyperparameter search space for a decision tree.}
\begin{center}
\footnotesize
%\begin{adjustbox}{width=\textwidth}{
\begin{tabular}{@{}llcc@{}}
 \toprule
 \textbf{Hyperparameter} & \textbf{Type} & \textbf{Range} & \textbf{Log scale} \\
 \midrule
  $criterion$ & Categorical & $\{Gini, Entropy\}$ &  -\\ \midrule
 $max\_depth$ & Integer & $[1, 21]$ & -  \\ \midrule
 $min\_samples\_split$ & Integer & $[2, 11]$ & -  \\\midrule
 $max\_leaf\_nodes$ & Integer & $[3, 26]$ & -  \\\midrule
 $splitter$ & Categorical & $\{Best, Random\}$ &  -\\  \bottomrule 
\end{tabular}
%}\end{adjustbox}
\end{center}

\end{table}

%% file: Tables/catboost_search_space.tex
\begin{table}[ht]
\caption{\label{app:catboost_search_space} The hyperparameter search space for CatBoost.}
\begin{center}
\footnotesize
\resizebox{0.6\linewidth}{!}{
%\begin{adjustbox}{width=\textwidth}{
\begin{tabular}{@{}llcc@{}}
 \toprule
 \textbf{Hyperparameter} & \textbf{Type} & \textbf{Range} & \textbf{Log scale} \\
 \midrule
 $learning\_rate$ & Continuous & $[1e-5, 1]$ & \checkmark  \\ \midrule
 $random\_strength$ & Integer & $[1, 20]$ & -  \\ \midrule
 $l2\_leaf\_reg$ & Continuous & $[1, 10]$ & \checkmark \\ \midrule
 $bagging\_temperature$ & Continuous & $[1e-6, 1]$ & \checkmark \\ \midrule
 $leaf\_estimation\_iterations$ & Integer & $[1, 20]$ & -  \\ \midrule
 $iterations$ & Integer & $[100, 4000]$ & -
  \\ \bottomrule 
\end{tabular}}
%}\end{adjustbox}
\end{center}

\end{table}

%% file: Tables/random_forest_search_space.tex
\begin{table}[ht]
\caption{\label{app:random_forest_search_space} The hyperparameter search space for Random Forest.}
\begin{center}
\footnotesize
\resizebox{0.6\linewidth}{!}{
%\begin{adjustbox}{width=\textwidth}{
\begin{tabular}{@{}llcc@{}}
 \toprule
 \textbf{Hyperparameter} & \textbf{Type} & \textbf{Range} & \textbf{Log scale} \\
 \midrule
 $criterion$ & Categorical & $\{Gini, Entropy\}$ &  -\\ \midrule
 $max\_depth$ & Integer & $[1, 21]$ & -  \\ \midrule
 $min\_samples\_split$ & Integer & $[2, 11]$ & -  \\\midrule
 $max\_leaf\_nodes$ & Integer & $[3, 26]$ & -  \\\midrule
 $n\_estimators$ & Integer & $[100, 4000]$ &  -\\  \bottomrule 
\end{tabular}}
%}\end{adjustbox}
\end{center}

\end{table}

%% file: Tables/tabnet_search_space.tex
\begin{table}[!ht]
\centering
\footnotesize
\caption{\label{app:tabnet_search_space}Hyperparameter search space for the TabNet model.}
\begin{tabular}{@{}llcc@{}}
\toprule
\textbf{Hyperparameter}         & \textbf{Type}        & \textbf{Range}         &      \textbf{Log scale}         \\ \midrule
$n\_a$                       & Categorical          & $\{8, 16, 24, 32, 64, 128\}$    &  -          \\ \midrule
$learning\_rate$             & Categorical          & $\{0.005, 0.01, 0.02, 0.025\}$      &  -      \\ \midrule
$gamma$                      & Categorical          & $\{1.0, 1.2, 1.5, 2.0\}$           &  -       \\ \midrule
$n\_steps$                   & Categorical          & $\{3, 4, 5, 6, 7, 8, 9, 10\}$      &  -       \\ \midrule
$lambda\_sparse$             & Categorical          & $\{0, 0.000001, 0.0001, 0.001, 0.01, 0.1\}$ &  -\\ \midrule
$batch\_size$                & Categorical          & $\{256, 512, 1024, 2048, 4096, 8192, 16384, 32768\}$ &  -\\ \midrule
$virtual\_batch\_size$       & Categorical          & $\{256, 512, 1024, 2048, 4096\}$    &  -      \\ \midrule
$decay\_rate$                & Categorical          & $\{0.4, 0.8, 0.9, 0.95\} $          &  -      \\ \midrule
$decay\_iterations$          & Categorical          & $\{500, 2000, 8000, 10000, 20000\} $  &  -     \\ \midrule
$momentum$                   & Categorical          & $\{0.6, 0.7, 0.8, 0.9, 0.95, 0.98\} $ &  -     \\ \bottomrule
\end{tabular}
\end{table}